\documentclass{article}

\usepackage{arxiv}

\usepackage[utf8]{inputenc} 
\usepackage[T1]{fontenc}    
\usepackage{hyperref}       
\usepackage{url}            
\usepackage{booktabs}       
\usepackage{amsfonts}       
\usepackage{nicefrac}       
\usepackage{microtype}      
\usepackage{lipsum}		
\usepackage{graphicx}
\usepackage{natbib}
\usepackage{doi}

\usepackage{microtype}
\usepackage{subcaption}
\usepackage{graphicx}
\usepackage{booktabs} 
\usepackage{multirow}

\usepackage{hyperref}

\usepackage{float}

\usepackage[dvipsnames, svgnames, x11names]{xcolor}
\usepackage{dsfont}
\usepackage{setspace}

\usepackage[algo2e,ruled,linesnumbered,boxed]{algorithm2e}
\usepackage{algorithmic}
\usepackage{algorithm}
\SetKw{myInput}{Input}
\SetKw{myOutput}{Output}

\usepackage{amsmath}
\usepackage{amssymb}
\usepackage{mathtools}
\usepackage{amsthm}
\usepackage{amsfonts}       
\usepackage{nicefrac}       
\usepackage{microtype}      
\usepackage{times}
\usepackage{epsfig}
\usepackage{float}
\usepackage{amsmath}
\allowdisplaybreaks[4]
\usepackage{amssymb}
\usepackage{appendix}
\usepackage{amsthm}
\usepackage{rotating}
\usepackage{autobreak}

\usepackage[capitalize,noabbrev]{cleveref}

\theoremstyle{plain}

\theoremstyle{definition}

\theoremstyle{remark}

\usepackage[textsize=tiny]{todonotes}

\title{Structured Cooperative Learning with\\Graphical Model Priors}

\author{%
Shuangtong Li$^{1}$ \quad Tianyi Zhou$^{2}$ \quad Xinmei Tian$^{1\  3}$ \quad Dacheng Tao$^4$ \\
$^1$University of Science and Technology of China \quad $^2$University of Maryland, College Park \\
\quad $^3$Institute of Artificial Intelligence, Hefei Comprehensive National Science Center \quad $^4$The University of Sydney\\
\texttt{lst2015@mail.ustc.edu.cn} \quad
\texttt{tianyi@umd.edu} \\
\texttt{xinmei@ustc.edu.cn} \quad 
\texttt{dacheng.tao@gmail.com}
}

\renewcommand{\shorttitle}{Structured Cooperative Learning}

\hypersetup{
pdftitle={A template for the arxiv style},
pdfsubject={q-bio.NC, q-bio.QM},
pdfauthor={David S.~Hippocampus, Elias D.~Striatum},
pdfkeywords={First keyword, Second keyword, More},
}

\begin{document}
\maketitle

\begin{abstract}
We study how to train personalized models for different tasks on decentralized devices with limited local data. We propose ``\underline{\textbf{S}}tructured \underline{\textbf{Coo}}perative \underline{\textbf{L}}earning (\textbf{SCooL})'', in which a cooperation graph across devices is generated by a graphical model prior to automatically coordinate mutual learning between devices. By choosing graphical models enforcing different structures, we can derive a rich class of existing and novel decentralized learning algorithms via variational inference. In particular, we show three instantiations of SCooL that adopt Dirac distribution, stochastic block model (SBM), and attention as the prior generating cooperation graphs. 
These EM-type algorithms alternate between updating the cooperation graph and cooperative learning of local models. They can automatically capture the cross-task correlations among devices by only monitoring their model updating in order to optimize the cooperation graph. 
We evaluate SCooL and compare it with existing decentralized learning methods on an extensive set of benchmarks, on which SCooL always achieves the highest accuracy of personalized models and significantly outperforms other baselines on communication efficiency. Our code is available at \href{https://github.com/ShuangtongLi/SCooL}{https://github.com/ShuangtongLi/SCooL}.
\end{abstract}


\section{Introduction}

  \begin{figure}[!h]
  \centering
			\includegraphics[width=0.7\linewidth]{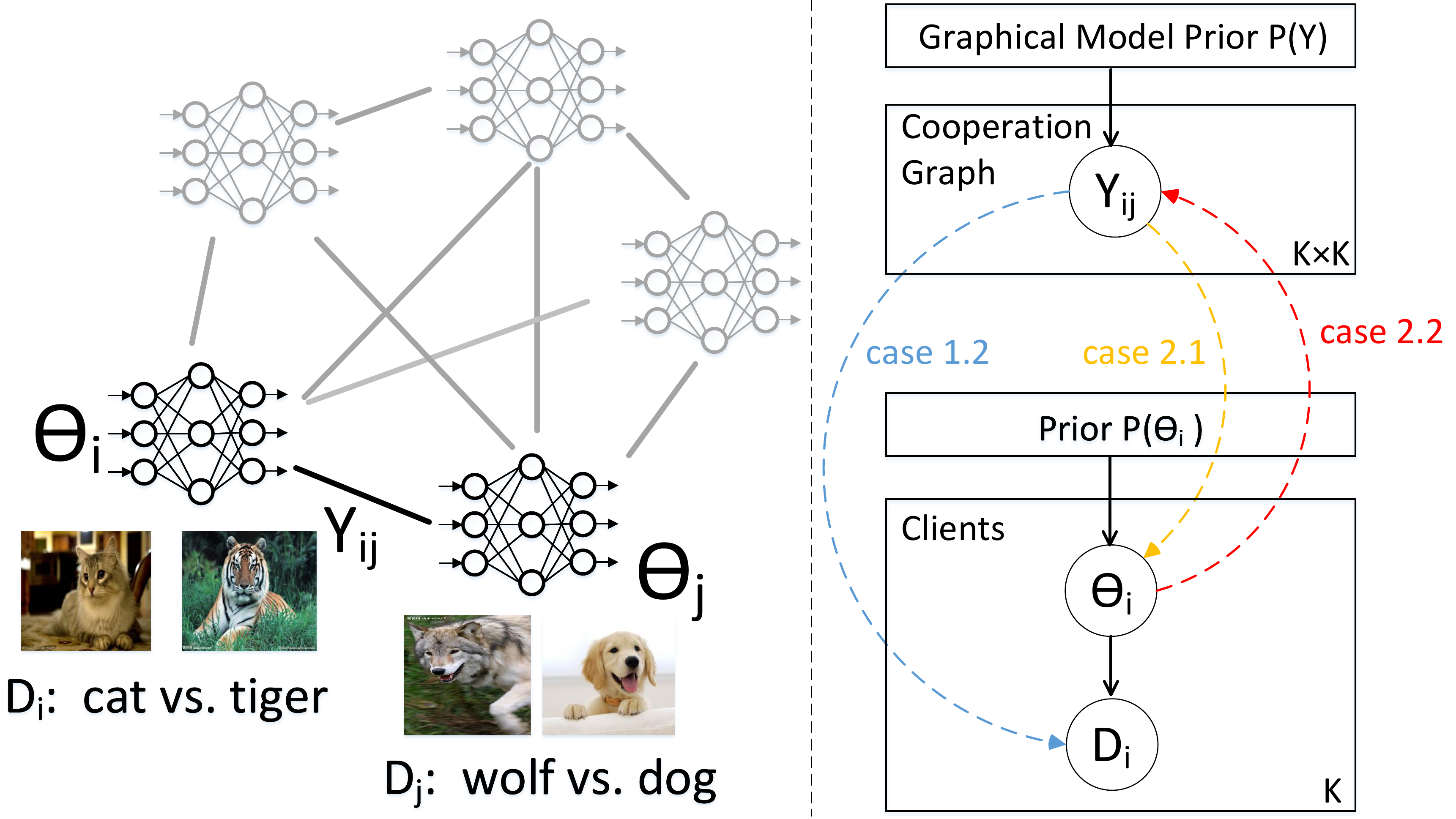}
		\caption{\footnotesize \textit{SCooL} framework. \textbf{Left}: \textit{SCooL} optimizes a $K\times K$ cooperation graph $Y$ together with $K$ personalized models $\theta_{1:K}$ for $K$ tasks, which transfer knowledge across the $K$ clients via decentralized learning. \textbf{Right}: \textit{SCooL}'s probabilistic model. $Y$ is generated from a graphical model prior while $\theta_i$ or data $D_i$ per client is generated based on $Y$. We discuss configurations (cases in Section~\ref{sec:configs}) of the probabilistic model and derive EM-type algorithms (Section~\ref{sec:em}) that alternately updates $Y$ and $\theta_{1:K}$. \looseness-1
        } 		
  \label{fig: SCooL framework}
	\end{figure}

Decentralized learning of personalized models (\textbf{DLPM}) is an emerging problem in a broad range of applications, in which multiple clients target different yet relevant tasks but no central server is available to coordinate or align their learning. A practical challenge is that each single client may not have sufficient data to train a model for its own task and thus has to cooperate with others by sharing knowledge, i.e., through cooperative learning. 
However, it is usually difficult for a client in the decentralized learning setting to decide when to cooperate with which clients in order to achieve the greatest improvement on its own task, especially when the personal tasks and local data cannot be shared across clients. Moreover, frequently communicating with all other clients is usually inefficient or infeasible. Hence, it is critical to find a sparse cooperation graph only relating clients whose cooperative learning is able to bring critical  improvement to their personalization performance. Since the local models are kept being updated, it is also necessary to accordingly adjust the graph to be adaptive to such changes in the training process.

Structural learning of a cooperation graph on the fly with decentralized learning of local models is an open challenge and can be prone to high variance caused by client heterogeneity and local data deficiency. Inspired Bayesian methods and their priors, we propose ``\underline{\textbf{S}}tructured \underline{\textbf{Coo}}perative \underline{\textbf{L}}earning (\textbf{SCooL})''. SCooL applies a probabilistic graphical model (PGM) as a structured prior enforcing certain structures such as clusters when generating the cooperation graph. By combining such a graphical model prior with the expressive power of neural networks on learning local tasks, we are able to develop a general framework for DLPM, from which we can derive a rich class of novel algorithms associated with different structured priors. In particular, we propose a probabilistic model to generate the cooperation graph and local models (Fig.~\ref{fig: SCooL framework} Right). Variational inference on this probabilistic model produces an approximate Maximum-A-Posteriori (MAP) estimation, which leads to an EM-type algorithm that alternately updates the cooperation graph and local models (Fig.~\ref{fig: SCooL framework} Left).  

We discuss several designs or configurations of the key components in the generative model and a general variational inference framework to derive EM algorithms for the model. For instance, we apply three different graphical model priors to generate the cooperation graph in the model and follow SCooL framework to derive three decentralized learning algorithms. While the Dirac Delta prior leads to an existing algorithm, i.e., D-PSGD~\cite{NIPS2017_f7552665}, the other two priors, i.e., stochastic block model (SBM) and attention, lead to two novel algorithms (\textit{SCooL-SBM} and \textit{SCooL-Attention}) that assume different structures and correlation among the local tasks. These two structural priors accelerate the convergence to a sparse cooperation graph (Fig.~\ref{fig:SCooL-graph-SBM}-\ref{fig:SCooL-graph-nonIID}), which can accurately identify the relevant tasks/clients and significantly save the communication cost (Fig.~\ref{fig:communication budget}). \looseness-1

In experiments on several decentralized learning benchmarks created from three datasets using two different schemes to draw non-IID tasks, SCooL outperforms SOTA decentralized and federated learning approaches on personalization performance (Table~\ref{tab:main result}) and computational/communication efficiency (Fig.~\ref{fig:test acc convergence}). We further investigate the capability of SCooL on recovering the cooperation graph pre-defined to draw non-IID tasks. The results explain how SCooL captures the task correlations to coordinate cooperation among relevant tasks and improve their own personalization performance. \looseness-1

 \section{Related work}
    \paragraph{Federated learning (FL)}~\cite{mcmahan2017communication}   Both empirical~\cite{hsieh2020non} and theoretical~\cite{karimireddy2020scaffold} studies find that the performance of FL degrades in non-IID settings when the data distributions (e.g., tasks) over devices are heterogeneous. Several strategies have been studied to address the non-IID challenge: modifying the model aggregation~\cite{NEURIPS2020_18df51b9,pmlr-v139-fraboni21a,chen2021fedbe,Wang2020Federated,balakrishnan2022diverse}, regularizing the local objectives with proximal terms~\cite{acar2021federated,li2018federated}. 
    or alleviating catastrophic forgetting in local training~\cite{xu2022acceleration}. These methods focus on improving the global model training to be more robust to non-IID distributions they can be sub-optimal for training personalized models for local tasks. Recent works study to improve the personalization performance in non-IID FL via: (1) trading-off between the global model and local personalization~\cite{li2021ditto,NEURIPS2020_f4f1f13c}; (2) clustering of local clients~\cite{sattler2020clustered, ghosh2020efficient,xie2021multi,Multi-Center_Federated_Learning}; (3) personalizing some layers of local models~\cite{li2021fedbn,liang2020think,pmlr-v139-collins21a,oh2022fedbabu,DualPersonalization}; (4) knowledge distillation~\cite{zhu2021data,afonin2022towards}; (5) training the global model as an initialization~\cite{fallah2020personalized} or a generator~\cite{pmlr-v139-shamsian21a} of local models; (6) using personalized prototypes~\cite{tan2022fedproto,tan2022federated}; (7) masking local updates~\cite{DBLP:conf/icml/Dai0H0T22}; or (8) learning the collaboration graph~\cite{Personalized_Federated_Learning_With_Structural_Information}. Most of them focus on adjusting the interactions between the global model and local personalized models. In contrast to DLPM (our problem), FL assumes a global server so direct communication among personalized models is not fully explored. Although clustering structures have be studied for FL, structure priors of cross-client cooperation graphs has not been thoroughly discussed.\looseness-1

	\paragraph{Decentralized learning (DL)} Earlier works in this field combine the gossip-averaging \cite{blot2016gossip} with SGD. Under topology assumptions such as doubly stochastic mixing-weights~\cite{NIPS2017_a74c3bae}, all local models can be proved to converge to a ``consensus model''~\cite{NIPS2017_f7552665} after iterating peer-to-peer communication. Although they show promising performance in the IID setting, \cite{hsieh2020non} points out that they suffer from severe performance degeneration in non-IID settings. 
	To tackle this problem, recent works attemp to improve the model update schemes or model structures, e.g., modifying the SGD momentum term~\cite{lin2021quasi}, replacing batch normalization with layer normalization~ \cite{hsieh2020non}, updating on clustered local models~\cite{khawatmi2017decentralized}, or modifying model update direction for personalized tasks~\cite{esfandiari2021cross}. Another line of works directly studies the effects of communication topology on consensus rates~\cite{DBLP:conf/icml/HuangSZYX22,yuan2022revisiting,song2022communicationefficient,vogels2022beyond}. Comparing to DLPM, these methods still focus on achieving a global consensus model rather than optimizing personalized models for local tasks. In addition, comparing to the cooperation graph in SCooL, their mixing weights are usually pre-defined instead of automatically optimized for local tasks. SPDB~\cite{lu2022decentralized} learns a shared backbone with personalized heads for local tasks. However, sharing the same backbone across all tasks might be sub-optimal and they do not optimize the mixing weights for peer-to-peer cooperation. 

\section{Probabilistic Cooperative Learning}

	\subsection{Probabilistic Modeling with Cooperation Graph}
	We study a probabilistic model whose posterior probability of local models $\theta_{1:K}$ given local data $D_{1:K}$ is defined by
 
	\begin{equation}
		\begin{aligned}\label{equ: template formulation}
			P(\theta_{1:K}|D_{1:K}) \propto P(\theta_{1:K},D_{1:K}) 
   = \int P(D_{1:K}|\theta_{1:K},Y)P(\theta_{1:K},Y)dY.
		\end{aligned}
  \end{equation}
    The cooperative learning of $\theta_{1:K}$ aims to maximize the posterior. Different from conventional decentralized learning methods like D-PSGD which fixes the cooperation graph or mixing weights, we explicitly optimize the cooperation graph $Y$ for more effective cooperation among decentralized models maximizing $P(D_{1:K}|\theta_{1:K},Y)$. 
The posterior in Eq.~\eqref{equ: template formulation} is decomposed into two parts: 
 the joint prior of $\theta_{1:K}$ and $Y$, and the joint likelihood of $D_{1:K}$ given $\theta_{1:K}$ and $Y$. By assuming different structures of the two parts (case 1.1-1.2 and case 2.1-2.3) and applying different priors $P(Y)$ for $Y$, we achieve a general framework from which we can derive a rich class of decentralized cooperative learning algorithms.\looseness-1

 \subsection{Configurations of Joint Likelihood and Prior}
 \label{sec:configs}
 In the following, we will discuss several possible configurations of the general probabilistic model in Eq.~\eqref{equ: template formulation}.

     \paragraph{Joint Likelihood $P(D_{1:K}|\theta_{1:K},Y)$}
    Maximizing this joint likelihood optimizes both models $\theta$ and the cooperation graph $Y$ to fit the datasets $D_{1:K}$. In a trivial case with $Y$ fixed, it reduces to classical decentralized learning. In contrast, the joint likelihood allows us to choose a cooperation graph $Y$ determining the data distributions of clients:

    \paragraph{case 1.1 $P(D_{1:K}|\theta_{1:K}):$} when $Y$ is pre-defined without affecting the data distribution, the joint likelihood can be designed as a simple product of likelihoods over all clients. 
     \begin{equation}
         P(D_{1:K}|\theta_{1:K}) = \prod_{i = 1}^K P(D_{i}|\theta_{i})
     \end{equation}
     \paragraph{case 1.2 $P(D_{1:K}|\theta_{1:K}, Y)$} enables us to optimize the cooperation graph to coordinate the training of multiple local models. For example, the following joint likelihood model leads to a multi-task learning objective:
		\begin{align}\label{equ: prob_D_given_Y}
			&P(D_{1:K}|\theta_{1:K},Y)=
			\prod_{i=1}^K P(D_{1:K}|\theta_i,Y)\nonumber = \prod_{i=1}^k{\bigg(P(D_i|\theta_i)\prod_{j\neq i ,Y_{ij}=1}}{P(D_j|\theta_i)}\bigg).
		\end{align}
  This objective leads to a personalized model $\theta_i$ learned from multiple sources of data $D_{1:K}$ and $Y$ provides the mixing weights for different sources: $Y_{ij}=1$ encourage a cooperation between client-$i$ and client-$j$ so the learning of $\theta_i$ can benefit from learning an additional task on $D_j$;  
  while $Y_{ij}=0$ indicates that learning task-$j$'s data $D_j$ hardly bring improvement to $\theta_i$ on task-$i$. 
  
     \paragraph{Joint Priors of Personalized Models and Cooperation Graph $P(\theta_{1:K},Y)$} can be parameterized in three different forms by presuming different dependencies of models and cooperation graphs:
     \begin{equation}
         \notag \begin{aligned}
              &\textbf{case 2.1}: P(\theta_{1:K}|Y)P(Y), \ \ \ \ \text{$\theta_{1:K}$ is derived from $Y$}.  \\
              &\textbf{case 2.2}: P(Y|\theta_{1:K})P(\theta_{1:K}), \text{$\theta_{1:K}$ determines $Y$}.  \\
              &\textbf{case 2.3}: P(\theta_{1:K})P(Y), \ \ \ \ \ \ \ \ \ \text{$\theta_{1:K}$ is independent to $Y$}.
         \end{aligned}
     \end{equation}
    By choosing a joint prior from \textbf{case 2.1-2.3} and combining it with a joint likelihood chosen from \textbf{case 1.1-1.2}, we are able to create a rich family of probabilistic models that relate local models through their cooperation graph. In particular, the cooperation graph $Y$ can guide the cross-client cooperation by relating either different clients' data (case 1.2) or their models (case 2.1-2.2). Practical designs of likelihood and prior need to consider the feasibility and efficiency of inference.  
    
    The generation of cooperation graph $Y$ in the probabilistic model plays an important role in determining knowledge transfer across clients in cooperative learning. As shown in Section~\ref{sec:examples}, if clients' tasks have a clustering structure and the clients belonging to the same cluster have a higher probability to cooperate, we can choose a stochastic block model (SBM) $P(Y)$ as the prior to generating $Y$; if we encourage the cooperation between clients with similar tasks or models, we can generate $Y_{ij}$ via $P(Y_{ij}|\theta_i, \theta_j)$ according to the similarity between $\theta_i$ and $\theta_j$, which can be captured by an ``attention prior'' that will be introduced later.

\subsection{Variational Inference of Cooperation Graph \& Cooperative Learning of Personalized Models}
\label{sec:em}
    Maximizing the posterior in Eq.~\eqref{equ: template formulation} requires an integral on latent variable $Y$, which may not have a closed form or is expensive to compute by sampling methods.  
    Hence, we choose to use variational inference with mean field approximation~\cite{DBLP:journals/ml/JordanGJS99} to derive an EM algorithm that alternately updates the cooperation graph and local models using efficient closed-form updating rules. Despite possible differences in the concrete forms of likelihood and prior, the derived EM algorithm for different probabilistic models shares the same general form below.\looseness-1
    
    For observations $X$ (e.g., $X=D_{1:K}$), the set of all latent variables $Z$ (e.g., $Z\supseteq Y$), and the set of all model parameters $\Phi$ (e.g., $\Phi\supseteq \theta_{1:K}$), the posterior is lower bounded by
	\begin{equation}
		\begin{aligned}
			\log p(X|\Phi) = \log \int p(X, Z|\Phi)dZ 
			\ge \int q(Z)\log \frac{p(X, Z|\Phi)}{q(Z)}dZ := H(q, \Phi).
		\end{aligned}
	\end{equation}
	EM algorithm aims to maximize $L(q, \Phi)$ by iterating between the following E-step and M-step:
		
  \textbf{E-step} finds distribution $q$ to maximize the lower bound:
		\begin{equation}
			q \leftarrow \mathop{\arg \max}_{q} H(q, \Phi).
		\end{equation}
  	However, directly optimizing $q$ is usually intractable. Hence, we resort to mean-field theory that approximates $q(Z)$ by a product distribution with variational parameter $\beta_i$ for each latent variable $Z_i$, i.e., 
	\begin{equation}
		\begin{aligned}
			q_{\beta}(Z) = \prod_i q_{i}(Z_i|\beta_i). 
		\end{aligned}
	\end{equation}
    Then the E-step reduces to:
	\begin{equation}
		\beta \leftarrow \mathop{\arg \max}_{\beta} H(q_{\beta}, \Phi).
	\end{equation}
    For SCooL, its E-step has a closed-form solution $F(\cdot)$ to the variational parameters $w$ of the cooperation graph $Y$ so its E-step has the form of
    \begin{equation}\label{equ:SCooL-E}
        \begin{aligned}
        &w_{ij} \leftarrow F\bigg(\log P(D_j|\theta_i), \beta,\Phi \bigg)~\forall i,j\in[K]. \\
        &\text{Update other variational parameters in }\{\beta\} \backslash \{w_{ij}\}.
        \end{aligned}
   \end{equation}
    \textbf{M-step} optimizes parameters $\Phi$ given the updated $q$, i.e.,
	\begin{equation}
			{\Phi} \leftarrow \mathop{\arg \max}_{\Phi} H(q, \Phi).
	\end{equation}
    For SCooL, its M-step applies gradient descent to optimize the local models $\theta_{1:K}$.    

     \begin{align}\label{equ:SCooL-M}
    &\theta_i \leftarrow \theta_i - \eta_1 \bigg( 
    \sum_{j\neq i}w_{ij}\nabla L(D_j;\theta_i)
 + \nabla L(D_i;\theta_i)+G(\beta,\Phi)\bigg). \nonumber\\
    &\text{Update other observable variables }\{\Phi\}\backslash \{\theta_{i}\},
     \end{align}
    where $L(D; \theta)$ is the loss of model $\theta$ computed on dataset $D$ and $G(\beta, \Phi)$ is a term that only depends on $\beta$ and $\Phi$.

\textbf{Remarks:} 
In E-step shown in Eq.~\eqref{equ:SCooL-E}, SCooL updates $w_{ij}$ based on the ``cross-client loss'' $\log P(D_j|\theta_i)$ that evaluates the personalized model $\theta_i$ of client-$i$ on the dataset $D_j$ of client-$j$. Intuitively, a higher log-likelihood $\log P(D_j|\theta_i)$ implies that the tasks on client-$i$ and client-$j$ are similar so $\theta_i$ can be improved by learning from $\theta_j$ via a larger $w_{ij}$ (or $Y_{ij}$) in the cooperation graph.
In M-Step shown in Eq.~\eqref{equ:SCooL-M}, SCooL trains each personalized model $\theta_i$ by not only using its own gradient $\nabla L(D_i;\theta_i)$ but also aggregating the gradients $\nabla L(D_j;\theta_i)$ computed on other clients with \textbf{mixing weights} $w_{ij}$ from the cooperation graph. This encourages cooperative learning among clients with similar tasks. In Appendix~\ref{sec:SCooL-implementation}, we discuss a practical approximation to $\nabla L(D_j;\theta_i)$ that avoids data sharing between clients and saves communication cost without degrading cooperative learning performance. 


Therefore, by iterating between E-step and M-step for SCooL, we optimize the cooperation graph to be adaptive to the local training progress on clients and their latest personalized models, which are then updated via cooperative learning among relevant clients on the optimized graph. \looseness-1

\section{Graphical Model Priors for Cooperation Graph \& Three Instantiations of SCooL}\label{sec:examples}
\begin{algorithm2e}[!h]
    	\setstretch{0.5}
		\caption{\textbf{S}tructured \textbf{Coo}perative \textbf{L}earning}
		\label{alg:SCooL algorithm}
			\myInput{ 
	$\{D_i\}_{i=1}^K, S, T$
 }\\
   \myOutput{$\theta_{1:K}$}\\
		Initialize personalized models $\theta_{1:K}$, latent variational parameters $\beta$, observable variables $\Phi$\\ 
			\For{$t=0\to T$}{
   \For{client $i=1\to K$ \textbf{in parallel}}{
       \emph{E-step}:\\
             $w_{ij} \leftarrow F\bigg(\log P(D_j|\theta_i), \beta,\Phi \bigg).$ \\
         Update $\{\beta\}\backslash \{w_{ij}\}$.\\
        Examples: \\
        \qquad \colorbox{Orange}{SCooL-SBM: \ Eq.~(\ref{equ:SBM w update})-(\ref{equ:SBM phi update})}\\
         \qquad \colorbox{pink}{SCooL-attention: Eq.~(\ref{equ:attention w update})}\\
        \emph{M-step}:\\
        \For{\text{local SGD step }$m=0:s$}{
          $\theta_i \leftarrow \theta_i - \eta_1 \bigg( \nabla L(D_i;\theta_i)+$\\
         $\sum_{j\neq i}w_{ij}\nabla L(D_j;\theta_i)+G(\beta,\Phi)\bigg)$.} 
        Update $\{\Phi\}\backslash \{\theta_i\}$.\\
         Examples: \\
         \qquad \colorbox{Orange}{SCooL-SBM: Eq.~(\ref{equ:SBM theta update})-(\ref{equ:SBM B update})}\\
         \qquad\colorbox{pink}{SCooL-attention: Eq.~(\ref{equ:attention theta update})-(\ref{equ:attention phi update})}\\
        }}
\end{algorithm2e}
\begin{figure}[htbp]
		\begin{center}
			\includegraphics[width=0.6\linewidth]{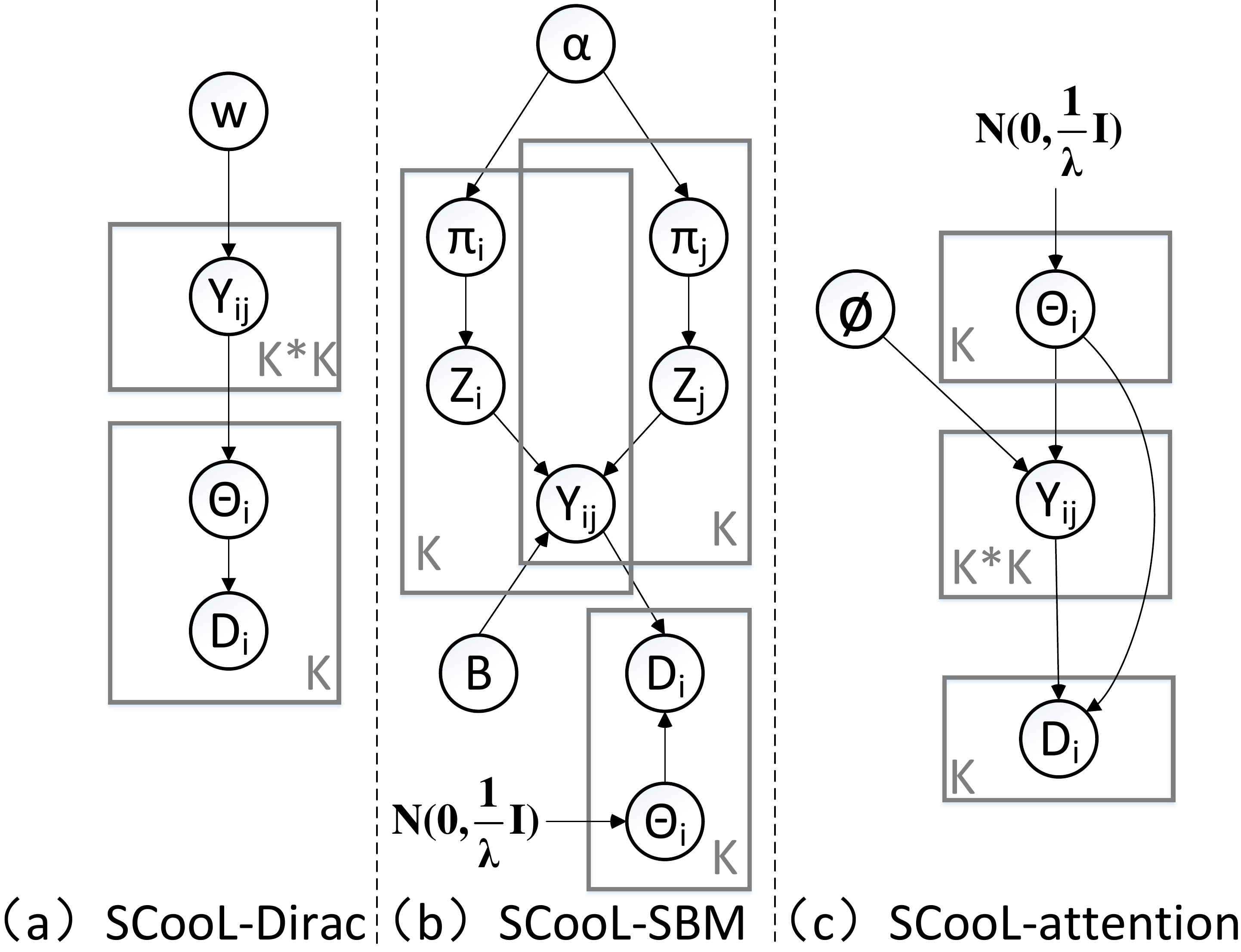}
		\end{center}

		\caption{\footnotesize 
			Probabilistic models used in SCooL examples.
		}
		\label{fig:all graphical models}
	\end{figure}
In this section, we derive three instantiations of SCooL algorithms associated with three graphical model priors $P(Y)$ used to generate the cooperation graph $Y$ in the probabilistic model. 
We first show that D-PSGD~\cite{NIPS2017_f7552665} can be derived as a special case of SCooL when applying a Dirac Delta prior to $Y$. We then derive SCooL-SBM and SCooL-attention that respectively use stochastic block model (SBM) and an ``attention prior'' to generate a structured $Y$. The probabilistic models for these three SCooL examples are shown in Fig.\ref{fig:all graphical models}.
	\subsection{Dirac Delta Prior Leads to D-PSGD}
	We choose prior $P(Y)$ as a simple Dirac Delta distribution and define $P(\theta|Y)$ as a manifold prior~\cite{belkin2006manifold} based on the pairwise distance between local models:
	\begin{align}
			&Y \sim \delta(w) \\
			&P(\theta_{1:K}|Y) \propto  \exp(-\frac{\lambda}{2}\sum_{1\leq i,j\leq K}Y_{ij}||\theta_i-\theta_j||^2) \\
			&P(D_{1:K}|\theta_{1:K}) \propto \prod_{i=1}^K P(D_{i}|\theta_{i}).
		\end{align}
	We then choose \textbf{case 1.1} as the likelihood and \textbf{case 2.1} as the prior. Hence, maximizing posterior or MAP is:
		\begin{align}
			\mathop{\arg \max}_{\theta_{1:K}} \ \ \ \log P(\theta_{1:K}|D_{1:K}) \nonumber 
			=\mathop{\arg \min}_{\theta_{1:K}} \ \ \ 
			\sum_{i=1}^K L(D_i;\theta_i)+
			\frac{\lambda}{2}\sum_{i,j}w_{ij}||\theta_i-\theta_j||^2.
		\end{align}
	The above MAP can be addressed by gradient descent:
	\begin{align}
			&\theta_i \leftarrow \theta_i -\alpha\bigg( \nabla_{\theta}L(D_i;\theta_i) + \frac{\lambda}{2}\sum_{j=1}^K(w_{ij}+w_{ji})(\theta_i-\theta_j)\bigg) \nonumber \\
			&\overset{\textcircled{1}}{=}\theta_i -\alpha\bigg( \nabla_{\theta}L(D_i;\theta_i) + \lambda\sum_{j=1}^K w_{ij}(\theta_i-\theta_j)\bigg) \nonumber \\
			&\overset{\textcircled{2}}{=}\theta_i -\alpha \nabla_{\theta}L(D_i;\theta) -\alpha \lambda \bigg(\theta_i-\sum_{j=1}^K w_{ij}\theta_j\bigg),
		\end{align}	
	where \textcircled{1} holds because we enforce $w_{ij}=w_{ji}$, and \textcircled{2} holds due to the constraint $\sum_{j=1}^K w_{ij}=1$.
	
	Taking $\lambda = \frac{1}{\alpha}$, we finally obtain the update rule as:
	\begin{equation}
		\theta_i \leftarrow \sum_{j=1}^K w_{ij}\theta_j-\alpha \nabla_{\theta}L(D_i;\theta_i)
	\end{equation}
	Hence, it exactly reconstruct the D-PSGD~\cite{NIPS2017_f7552665} algorithm (See Appendix~\Ref{sec:D-PSGD} for details of D-PSGD).

	\subsection{SCooL-SBM with Stochastic Block Model Prior}
	In cooperative learning, we can assume a clustering structure of clients such that the clients belonging to the same community benefit more from their cooperation. This structure can be captured by stochastic block model (SBM)~\cite{holland1983stochastic}, which is used as the prior generating the cooperation graph in SCooL. In particular, 
 \begin{itemize}
		\item For each client $i \in [K]$:
		\begin{itemize}
			\item Draw an $M$-dimensional membership probability distribution as a vector $\vec{\pi}_i \sim
			\textrm{Dirichlet}( \vec\alpha )$.
			\item Draw membership label
			$\vec z_{i} ~ \sim {\rm Multinomial}(\vec\pi_{i})$.
		\end{itemize}
		\item For each pair of clients $(i,j) \in [K] \times [K]$:
		\begin{itemize}
			\item Sample $Y_{ij} \sim {\rm
				Bernoulli}(\vec z_{i}^{\ T} B ~ \vec z_{j})$ that determines the cooperation between client pair $(i,j)$.
		\end{itemize}
	\end{itemize}
	Hence, the marginal distribution of $Y$ under SBM is:
	\begin{equation}
		\begin{aligned}
			&P(Y|\vec{\alpha},B)=\int P(Y,\vec{\pi}_{1:K},\vec{Z}_{1:K}|\vec{\alpha},B)d(\vec{\pi}_{1:K},\vec{Z}_{1:K}).\nonumber
		\end{aligned}
    \end{equation}
	We assume Gaussian priors for personalized models:
	\begin{equation}
		\begin{aligned}
			P(\theta_{1:K})\propto \exp(-\frac{\lambda}{2}\sum_{i}||\theta_i||^2).
		\end{aligned}
	\end{equation}
	

 \paragraph{\textbf{SCooL-SBM:}} 
 Since the generation of $Y$ does not depend on $\theta$, we can consider the joint prior in \textbf{case 2.3}. We further choose \textbf{case 1.2} as the likelihood. The EM algorithm for SCooL-SBM can be derived as the following (details are given in Appendix~\ref{sec:SCooL-SBM-EM}). 
 \begin{itemize}
		\item \textbf{E-step} updates $w$, $\gamma$, and $\Omega$, which are the variaiontal parameters of latent variables $Y$, $\pi$, and $z$, respectively. \looseness-1
			\begin{equation}\label{equ:SBM w update}
				w_{ij}\leftarrow \text{Sigmoid}\bigg(
				\log P(D_j|\theta_i)+\sum_{g,h} \vec{\Omega}_{ig}\vec{\Omega}_{jh} \log 
    B(g,h)
				-\sum_{g,h} \vec{\Omega}_{ig}\vec{\Omega}_{jh}\log (1-B(g,h))\bigg)
			\end{equation}	\begin{equation}\label{equ:SBM gamma update}
			\begin{aligned}
				\gamma_{ig} \leftarrow  \vec{\Omega}_{ig}
				+\vec{\alpha}_{g}
			\end{aligned}
  \end{equation}
		\begin{align}\label{equ:SBM phi update}
				\vec{\Omega}_{i\cdot}\leftarrow
				&\text{Softmax} \bigg(\sum_j w_{ij} \sum_h \vec\Omega_{jh} \log B(\cdot, h)+ 
			\sum_{j}w_{ji}\sum_h \vec\Omega_{jh} \log B(h, \cdot)
                +\psi(\gamma_{i\cdot})-\psi\left(\sum_g \gamma_{ig}\right)+ \\
				&\sum_j (1-w_{ij}) \sum_h \vec\Omega_{jh} \log (1-B(\cdot, h))+ +\sum_{j}(1-w_{ji})\sum_h \vec\Omega_{jh} \log (1-B(h, \cdot))\bigg)
			\end{align}
		\item \textbf{M-step} updates the model parameters $\theta_i$, $\vec{\alpha_{i}}$, and $B$.
		\begin{equation}\label{equ:SBM theta update}
			\begin{aligned}
				\theta_i \leftarrow \ \theta_i - \eta_1 \bigg(  
				\nabla_{\theta_i}L(D_i;\theta_i)+\sum_{j\neq i}w_{ij}\nabla_{\theta_i}L(D_j;\theta_i)
				+\lambda\theta_i
				\bigg)
			\end{aligned}
		\end{equation}
  \begin{equation}\label{equ:SBM alpha update}
			\begin{aligned}
				\vec{\alpha_{i}}\leftarrow \ 
				\vec{\alpha_{i}}+\eta_2\bigg(
				\sum_g \bigg(\psi(\gamma_{gi})-\psi(\sum_k \gamma_{gk})\bigg)-
				K\sum_g \psi(\vec{\alpha}_{g})
				+K
				\psi\left(\sum_k\vec{\alpha}_{k}\right)
				\bigg)
			\end{aligned}
		\end{equation}
  \begin{equation}\label{equ:SBM B update}
			\begin{aligned}
				B(i,j) \leftarrow \frac{\sum_{g,h}
					w_{gh}\vec{\Omega}_{gi}\vec{\Omega}_{hj}}{\sum_{g,h}
					\vec{\Omega}_{gi}\vec{\Omega}_{hj}}
			\end{aligned}
		\end{equation}
	\end{itemize}
	
	\subsection{SCooL-attention with Attention Prior}
 Instead of assuming a clustering structure, we can train an attention mechanism to determine whether every two clients can benefit from their cooperation. In the probabilistic model of SCooL, we can develop an attention prior to generate the cooperation graph. 
Specifically, we compute the attention between between $\theta_i$ and $\theta_j$ using a learnable metric $f(\cdot,\cdot)$, i.e.,
	\begin{equation}
p_{ij}=\frac{\exp(f(\theta_i, \theta_j))}{\sum_{l}\exp(f(\theta_i, \theta_\ell))},
	\end{equation}
In dot-product attention, $f(\theta_i, \theta_j)$ is the inner product between the representations of $\theta_i$ and $\theta_j$ produced by a learnable encoder $E(\cdot;\phi)$, e.g., the representation of $\theta_i$ is computed as $E(\theta_i^{t+\frac{1}{2}}-\theta_i^0;\phi)$. 
We compute the difference $\theta_i^{t+\frac{1}{2}}-\theta_i^0$ in order to weaken the impact of initialization $\theta_i^0$ and focus on the model update produced by gradient descent in the past training rounds. Hence, 
	\begin{equation}
		f(\theta_i, \theta_j)=\langle E(\theta_i^{t+\frac{1}{2}}-\theta_i^0;\phi), E(\theta_j^{t+\frac{1}{2}}-\theta_j^0;\phi)\rangle.
	\end{equation}
	Each row of $Y$ is a one-hot vector drawn from a categorical distribution defined by the attention scores between personalized models, i.e., 
	\begin{equation}
		\begin{aligned}
			\vec{Y_i} \sim \text{Categorical}(p_{i1},p_{i2},...,p_{iK}),~\forall~i\in [K]
		\end{aligned}
	\end{equation}
 In SCooL-attention's probabilistic model, we also use Gaussian as the prior for personalized models:
	\begin{equation}
		\begin{aligned}
			P(\theta_{1:K})\propto \exp(-\frac{\lambda}{2}\sum_{i}||\theta_i||^2),
		\end{aligned}
 \end{equation}
 \paragraph{\textbf{SCooL-attention}} 
 Hence, the above defines a joint prior in the form of \textbf{case 2.2}. We further adopt the likelihood in \textbf{case 1.2}. The EM algorithm for SCooL-attention can then be derived as the following (details given in Appendix~\ref{sec:SCooL-att-EM}). 
	\begin{itemize}
		\item \textbf{E-step} updates the variational parameters $w$ of cooperation graph $Y$, i.e., 
		\begin{equation}\label{equ:attention w update}
			\begin{aligned}
				w_{i\cdot}\leftarrow
				\text{Softmax}\bigg(
				\log P(D_\cdot|\theta_i)
				+\log p_{i\cdot}
				\bigg)
			\end{aligned}
		\end{equation}
\item \textbf{M-step} upates the model parameters $\theta_i$ and $\phi$, i.e.,	
\begin{align}\label{equ:attention theta update}
				\theta_i \leftarrow &\ \theta_i - \eta_1 \bigg(  
				\nabla_{\theta_i}L(D_i;\theta_i)+\sum_{j\neq i}w_{ij}\nabla_{\theta_i}L(D_j;\theta_i)+\lambda\theta_i
				-\sum_{ij}w_{ij}\nabla_{\theta_i}\log p_{ij}\bigg)
			\end{align}
\begin{align}\label{equ:attention phi update}
				&\phi \leftarrow \phi + \eta_2 \nabla_{\phi}
				\bigg(\sum_{ij}w_{ij}\log p_{ij}\bigg)
			\end{align}
	\end{itemize}

\section{Experiments}
 \begin{table}[htbp]
	\footnotesize
		\caption{\footnotesize The parameters of both non-IID~\cite{mcmahan2017communication} and non-IID SBM experimental setup.}
		\label{table:noniid}
		\begin{center}
			\begin{tabular}{lccccc}
			\toprule
				\bf Dataset &$M$ &$K$ &$N$ &model
				\\ \midrule
				CIFAR-10 &10 &100 &2 &two-layer CNN\\
				CIFAR-100 &100 &100 &10 &two-layer CNN\\
				MiniImageNet &100 &100 &10 &four-layer CNN\\
				\bottomrule			
			\end{tabular}
		\end{center}
	\end{table} 
	
\subsection{Experimental Setup} 
 To test the personalization performances of SCooL models, we draw classification tasks from two non-IID settings:
    \begin{itemize}
\item \textbf{Non-IID SBM}: a simpler non-IID setting. Given a dataset of $M$ classes, the totally $K$ clients can be divided into several groups. Clients within the same group are allocated with the same subset of $N$ classes, while clients from different groups do not have any shared classes. This task assignment distribution among clients can be described by a SBM model with an uniform membership prior $\alpha$.  
\item \textbf{Non-IID}~\cite{mcmahan2017communication}: A more challenging non-IID setting used in FedAvg~\cite{mcmahan2017communication}, where we randomly assign $N$ classes of data to every client from totally $M$ classes. Different from non-IID SBM setting, no pair of clients is ensured to share the same classes and data distribution.
    \end{itemize}
     We evaluate SCooL-SBM and SCooL-attention and compare them with several FL/DL baselines on three datasets: \textbf{CIFAR-10}~\cite{krizhevsky2009learning}, \textbf{CIFAR-100}, and \textbf{MiniImageNet} \cite{ravi2017optimization}, each having 50,000 training images and 10,000 test images. In Table~\ref{table:noniid}, we list the parameters for the two non-IID settings. Following the evaluation setting for personalized models in previous non-IID DL works~\cite{liang2020think, zhang2021personalized}, we evaluate each local model on all available test samples belonging to the classes in its local task. We run every experiment using five different random seeds and report their average test accuracy. We choose local models to be the two-layer CNN adopted in FedAvg~\cite{mcmahan2017communication} for CIFAR-10/100 and the four-layer CNN adopted in MAML~\cite{finn2017model} for MiniImageNet. Since Batch-Norm (BN) may have a detrimental effect on DL~\cite{hsieh2020non}, we replace all the BN layers~\cite{pmlr-v37-ioffe15} with group-norm layers~\cite{Wu_2018_ECCV}. The implementation details of SCooL-SBM and SCooL-attention are given in Appendix~\ref{sec:SCooL-implementation}.

	\paragraph{Baselines} 
	
	We compare our methods with a diverse set of baselines from federated learning (FL) and decentralized learning (DL) literature, as well as a \textbf{local SGD only baseline} without any model aggregation across clients. FL baselines include \textbf{FedAvg}~\cite{mcmahan2017communication} (the most widely studied FL method), \textbf{Ditto}~\cite{li2021ditto} achieving fairness and robustness via a trade-off between the global model and local objectives, and \textbf{FOMO}~\cite{zhang2021personalized} applying adaptive mixing weights to combine neighbors' models for updating personalized models. DL baselines include \textbf{D-PSGD}~\cite{NIPS2017_f7552665} with fixed mixing weights and topology, \textbf{CGA}~\cite{esfandiari2021cross} with a fixed topology but adaptive mixing weights for removing the conflict of cross-client gradients, \textbf{SPDB}~\cite{lu2022decentralized} with a shared backbone network but personalized heads for clients' local tasks, 
    \textbf{meta-L2C}~\cite{li2022learning} learning mixing weights to aggregate clients' gradients, 
    and \textbf{Dada}~\cite{zantedeschi2020fully} training local models with weighted regularization to the pairwise distance between local models.\looseness-1

    We run each baseline for 100 (communication) rounds or equally 500 local epochs if $>100$ rounds are needed, the same as our methods, except FedAvg which needs more (i.e., $>1000$) epochs to converge. For fair comparisons, we keep their communication cost per client and local epochs in each round to be no smaller than that of our methods. For FL baselines, the communication happens between the global server and clients, so we randomly select $10\%$ clients for aggregation and apply $5$ local epochs per client in each round. 
	For DL baselines, we let every client communicate with $\sim$10\% clients in every round. 
 we evaluate them in two settings, i.e., one local SGD step per round and $5$ local epochs per round. We evaluate each baseline DL method on multiple types of communication topology and report the best performance. More details are provided in Appendix~\ref{sec: decentralized details}. 
	\paragraph{Training hyperparameters}
	In all methods' local model training, we use SGD with learning rate of 0.01, weight decay of $5\times 10^{-4}$, and batch size of 10. 
	We follow the hyperparameter values proposed in the baselines' papers except the learning rate, which is a constant tuned/selected from $[0.01, 0.05, 0.1]$ for the best validation accuracy.

 	\begin{figure}[htbp]
  \centering
\begin{subfigure}
{0.49\linewidth}	\includegraphics[width=\linewidth]{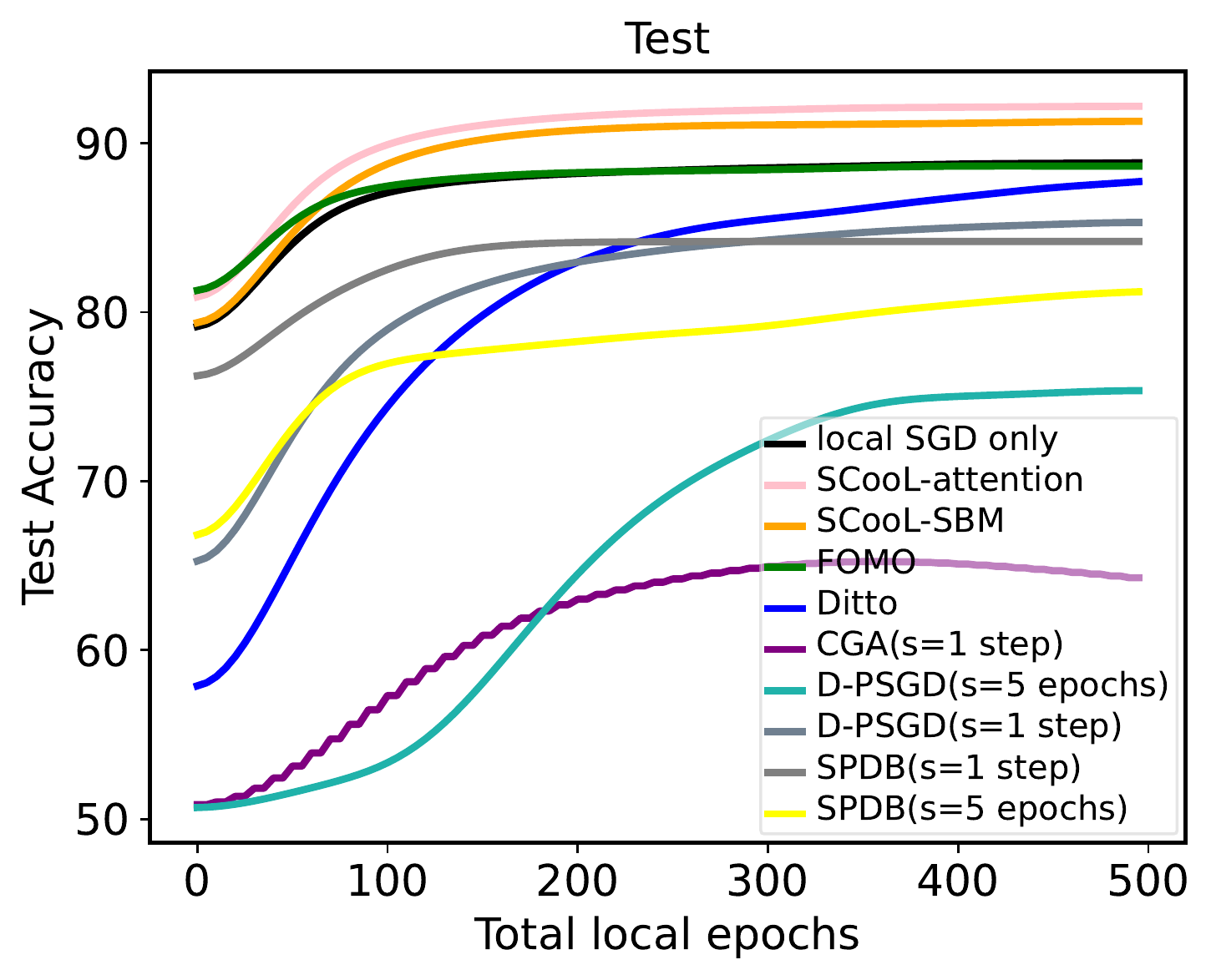}
 \end{subfigure}
 \begin{subfigure}
 {0.49\linewidth}	
\includegraphics[width=\linewidth]{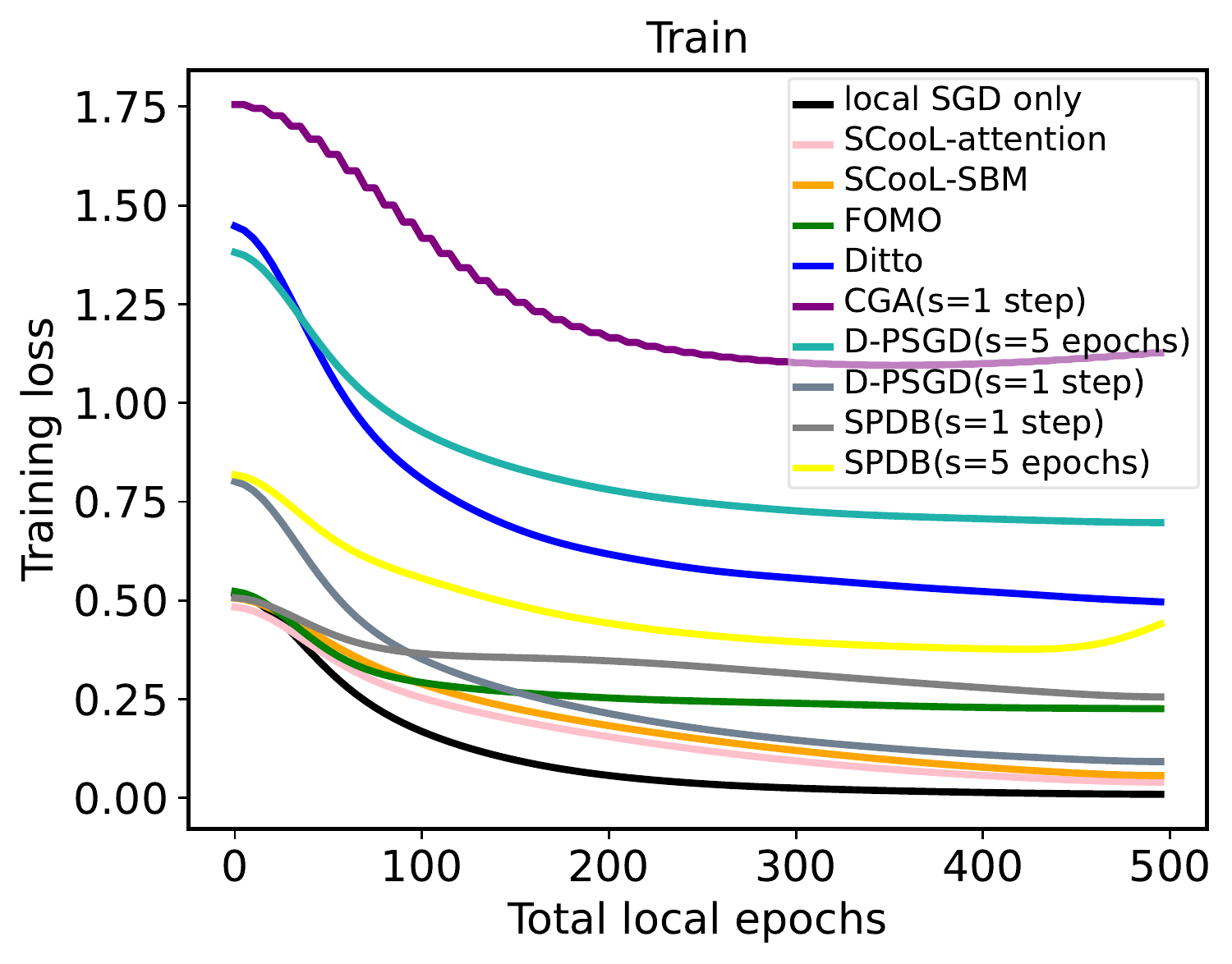}
 \end{subfigure}
		\caption{\footnotesize 
			\textbf{Test accuracy and training loss vs. total local epochs} 
on CIFAR-10. SCooL-SBM and SCooL-attention converge faster to better test/training 
performance than FL/DL baselines. 
FedAvg requires $>$ 1000 epochs to converge and is
not included.
		}
		\label{fig:test acc convergence}
	\end{figure}

\begin{figure}[!htbp]
\centering
		\begin{subfigure}[t]{\linewidth}
  \centering
		\includegraphics[width=0.75\linewidth]{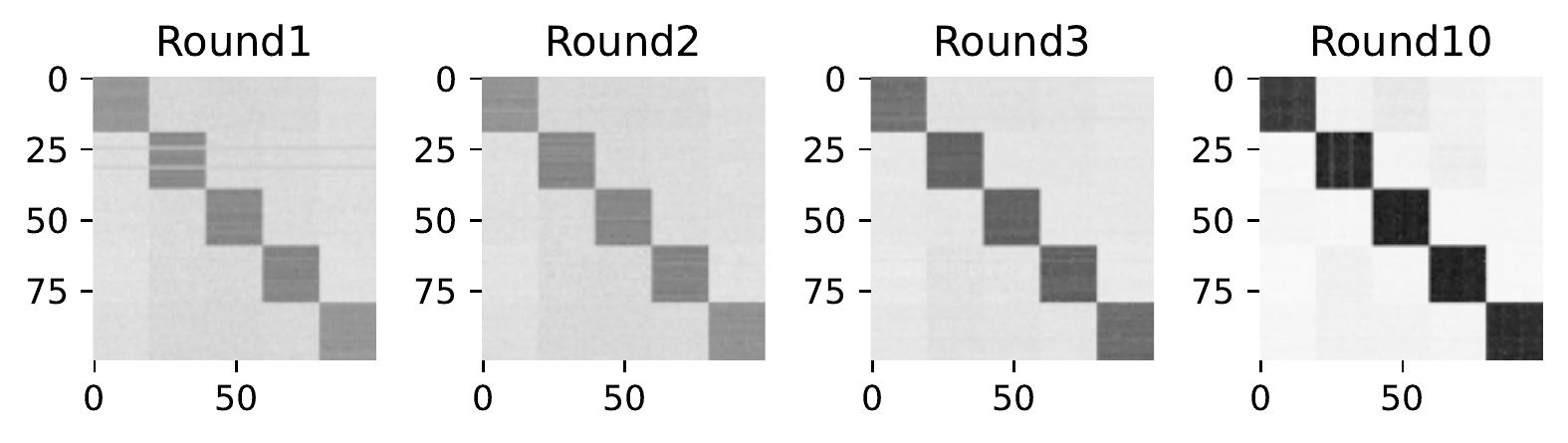}
			\caption{SCooL-SBM on non-IID SBM setting.}
			\label{fig:SCooL-SBM sibling mixing weights}
		\end{subfigure}%
       \quad
		\begin{subfigure}[t]{0.75\linewidth}
  \centering
			\includegraphics[width=\linewidth]{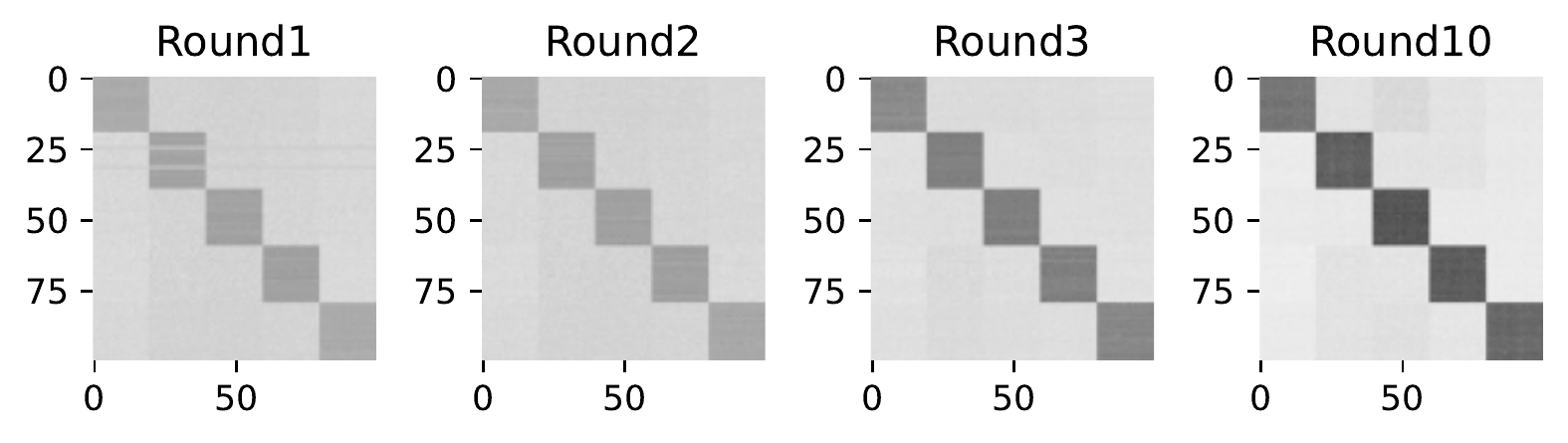}
    			\caption{SCooL-attention on non-IID SBM setting.}
			\label{fig:SCooL-attention sibling mixing weights}
		\end{subfigure}%
\caption{\footnotesize Cooperation graph weights $w$ by SCooL-SBM and SCooL-attention applied to the \textbf{non-IID SBM} setting. Both SCooL-SBM and SCooL-attention capture the ground-truth task similarity even in earlier training states, while the cooperation graphs of SCooL-SBM converge faster in non-IID SBM setting. \looseness-1}
        \label{fig:SCooL-graph-SBM}
\end{figure}

 \begin{table*}[!t]
  \centering
  \caption{Test accuracy (mean$\pm$std) of $100$ local models for non-IID tasks. SCooL-SBM and SCooL-attention outperform FL/DL baselines.}
\resizebox{\linewidth}{!}{%
    \begin{tabular}{l|l|ccc|ccc}
\toprule
         \multirow{2}[0]{*}{Methodology} & \multirow{2}[0]{*}{Algorithm}      & \multicolumn{3}{|c}{non-IID~\cite{mcmahan2017communication}} & \multicolumn{3}{|c}{non-IID SBM} \\
      		&&CIFAR-10 & CIFAR-100 & MiniImageNet & CIFAR-10 & CIFAR-100 & MiniImageNet \\
\hline
    		Local only & Local SGD only & 87.5$\pm$7.02  & 55.47$\pm$5.20 & 41.59$\pm$7.71 &   87.41$\pm$4.21    &   55.37$\pm$3.48    &  38.54$\pm$7.94 \\
\hline
    \multirow{3}[0]{*}{Federated} & FedAvg & 70.65$\pm$10.64 & 40.15$\pm$7.25 & 34.26$\pm$6.01 &   71.59$\pm$12.85   &  39.89$\pm$11.42     & 38.87$\pm$9.72 \\
          & FOMO  & 88.72$\pm$5.41 & 52.44$\pm$5.09 & 44.56$\pm$4.31 &  90.30$\pm$2.67    &    67.31$\pm$4.81   & 42.72$\pm$2.23 \\
          & Ditto & 87.32$\pm$6.42 & 54.28$\pm$5.31 & 42.73$\pm$5.19 &   88.13$\pm$7.43   &  54.34$\pm$5.42     & 42.16$\pm$5.46 \\
\hline
    \multirow{6}[0]{*}{Decentralized} & D-PSGD(s=1 step) & 83.01$\pm$7.34 & 40.56$\pm$6.94 & 30.26$\pm$5.75 &  85.20$\pm$4.05     &   48.15$\pm$4.77    &  37.43$\pm$3.59\\
          & D-PSGD(s=5 epochs) & 75.89$\pm$6.65 & 35.03$\pm$4.83 & 28.41$\pm$5.18 &  77.33$\pm$5.79     &   32.17$\pm$5.07    & 37.69$\pm$3.02 \\
          & CGA(s=1 step) & 65.65$\pm$12.66 & 30.81$\pm$10.79 & 27.65$\pm$11.78 &   69.93$\pm$5.34    &   36.91$\pm$7.58    &  25.54$\pm$1.95\\
          & CGA(s=5 epochs) & diverge & diverge & diverge &    diverge   &   diverge    & diverge \\
          & SPDB(s=1 step)  &  82.36$\pm$7.14   &    54.29$\pm$6.15           &   39.17$\pm$3.93   &   81.75$\pm$7.07    &   55.71$\pm$6.02   & 38.49$\pm$5.12\\
          & SPDB(s=5 epochs)  &  81.15$\pm$7.06   &   53.23$\pm$7.48   &    35.93$\pm$5.05  &   81.25$\pm$6.07    &  53.08$\pm$4.01     & 35.86$\pm$4.03 \\
          & Dada &
          85.65$\pm$6.36 &
          57.61$\pm$5.45 &
          37.81$\pm$7.15 &
          88.89$\pm$3.47 &
          64.62$\pm$4.77 &
          41.68$\pm$3.91 \\
          & meta-L2C &
          92.10$\pm$4.71 &
          58.28$\pm$3.09 &
          48.80$\pm$4.17 &
          91.84$\pm$2.40 &
          71.64$\pm$2.89 &
          49.95$\pm$1.97 \\
          
\hline
     \multirow{2}[0]{*}{SCooL(Ours)}     & SCooL-SBM & 91.37$\pm$5.03 & 58.76$\pm$4.30 & 48.69$\pm$5.21 &   \bf{94.14}$\pm$\bf{2.28}    &   \bf{72.27}$\pm$\bf{2.59}    & \bf{51.86}$\pm$\bf{1.64} \\
          & SCooL-attention & \bf{92.21}$\pm$\bf{5.15} & \bf{59.47}$\pm$\bf{4.95} & \bf{49.53}$\pm$\bf{3.29} & 93.98$\pm$3.85      &    72.03$\pm$2.71   & 51.69$\pm$2.80 \\
\bottomrule
    \end{tabular}%
}
  \label{tab:main result}%
\end{table*}%

    \subsection{Experimental Results}
    \paragraph{Test accuracy and convergence}  Table \ref{tab:main result} reports the test accuracy of all the 100 clients' models on their assigned non-IID tasks (mean$\pm$std over all clients). SCooL-SBM and SCooL-attention outperform FL/DL baselines by a large margin on all the three datasets. Moreover, SCooL-attention's prior can capture the pairwise similarity between clients in the non-IID setting so it outperforms SCooL-SBM. On the other hand, in the non-IID SBM setting, SCooL-SBM outperforms SCooL-attention since SCooL-SBM's prior is a better model of the SBM generated cooperation graph. In Fig.~\ref{fig:test acc convergence}, we compare the convergence of test accuracy and training loss for all methods in the non-IID setting, where SCooL-SBM and SCooL-attention converge faster than others. \looseness-1

   \begin{figure}[!tbp]
   \centering
		\begin{subfigure}{\linewidth}
  \centering
			\includegraphics[width=0.7\linewidth]{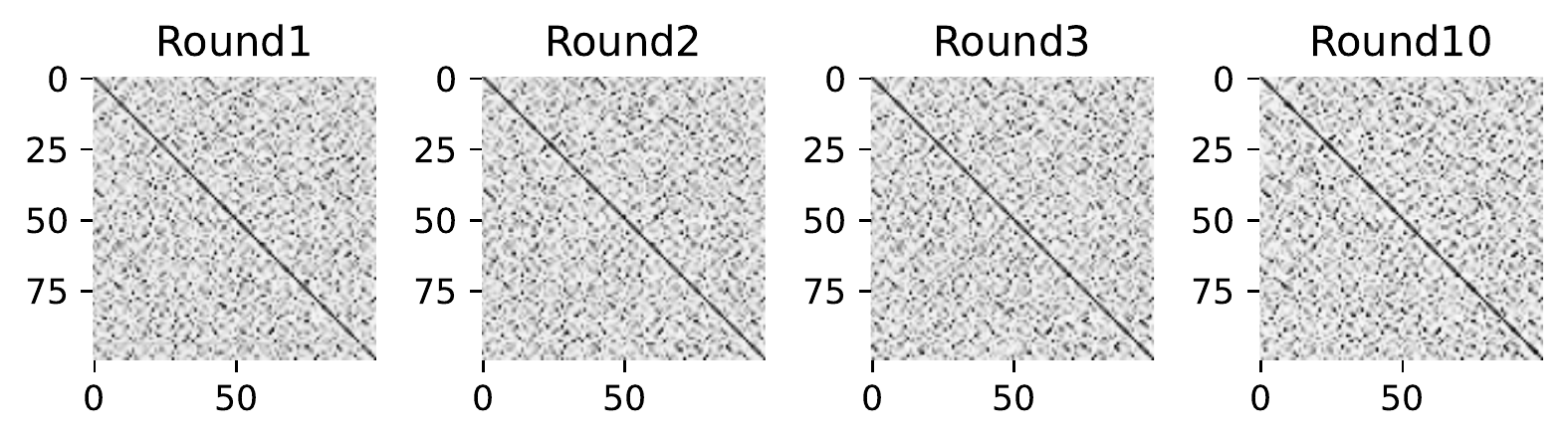}
			\caption{SCooL-SBM on non-IID~\cite{mcmahan2017communication} setting.}
			\label{fig:SCooL-SBM general mixing weights}
		\end{subfigure}%
       \quad
		\begin{subfigure}{0.7\linewidth}
  \centering
			\includegraphics[width=\linewidth]{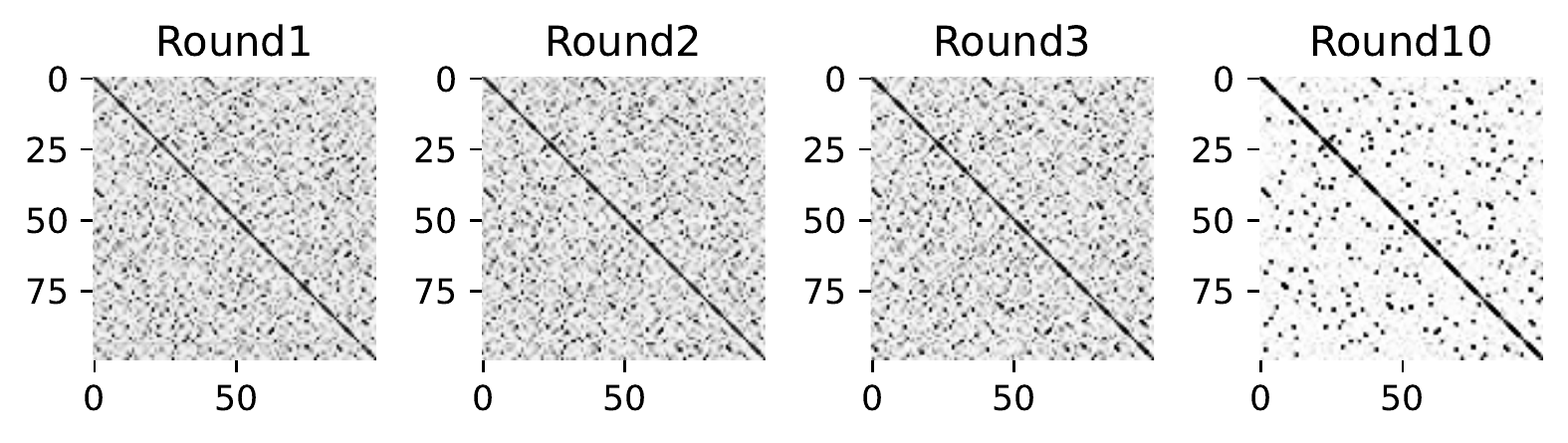}
    			\caption{SCooL-attention on non-IID~\cite{mcmahan2017communication} setting.}
			\label{fig:SCooL-attention general mixing weights}
		\end{subfigure}%
  \caption{\footnotesize Cooperation graph weights $w$ for 100 clients by SCooL-SBM and SCooL-attention applied to \textbf{non-IID} setting~\cite{mcmahan2017communication}. Both SCooL-SBM and SCooL-attention generate sparse cooperation graphs after a few rounds. SCooL-attention's cooperation graph converge faster than SCooL-SBM. \looseness-1} 
		\label{fig:SCooL-graph-nonIID}
	\end{figure}
        \paragraph{Learned cooperation graphs} In Fig. \ref{fig:SCooL-SBM sibling mixing weights}-\ref{fig:SCooL-attention sibling mixing weights}, 
        we report how the cooperation graphs produced by SCooL-SBM and SCooL-attention over communication rounds for non-IID SBM setting. Both methods capture the true task relationships after only a few training rounds. The faster convergence of SCooL-SBM indicates that SCooL-SBM's prior is a better model capturing the SBM cooperation graph structure. In Fig.~\ref{fig:SCooL-SBM general mixing weights}-\ref{fig:SCooL-attention general mixing weights}, we report the learned mixing weights in the non-IID setting. Both algorithms can quickly learn a sparse cooperation graph very early, which significantly reduces the communication cost for later-stage training. In this setting, SCooL-attention is better and faster than SCooL-SBM on capturing the peer-to-peer correlations. 

  \begin{figure}[!htbp]
	\centering	\begin{subfigure}{0.48\linewidth}
			\includegraphics[width=\linewidth]{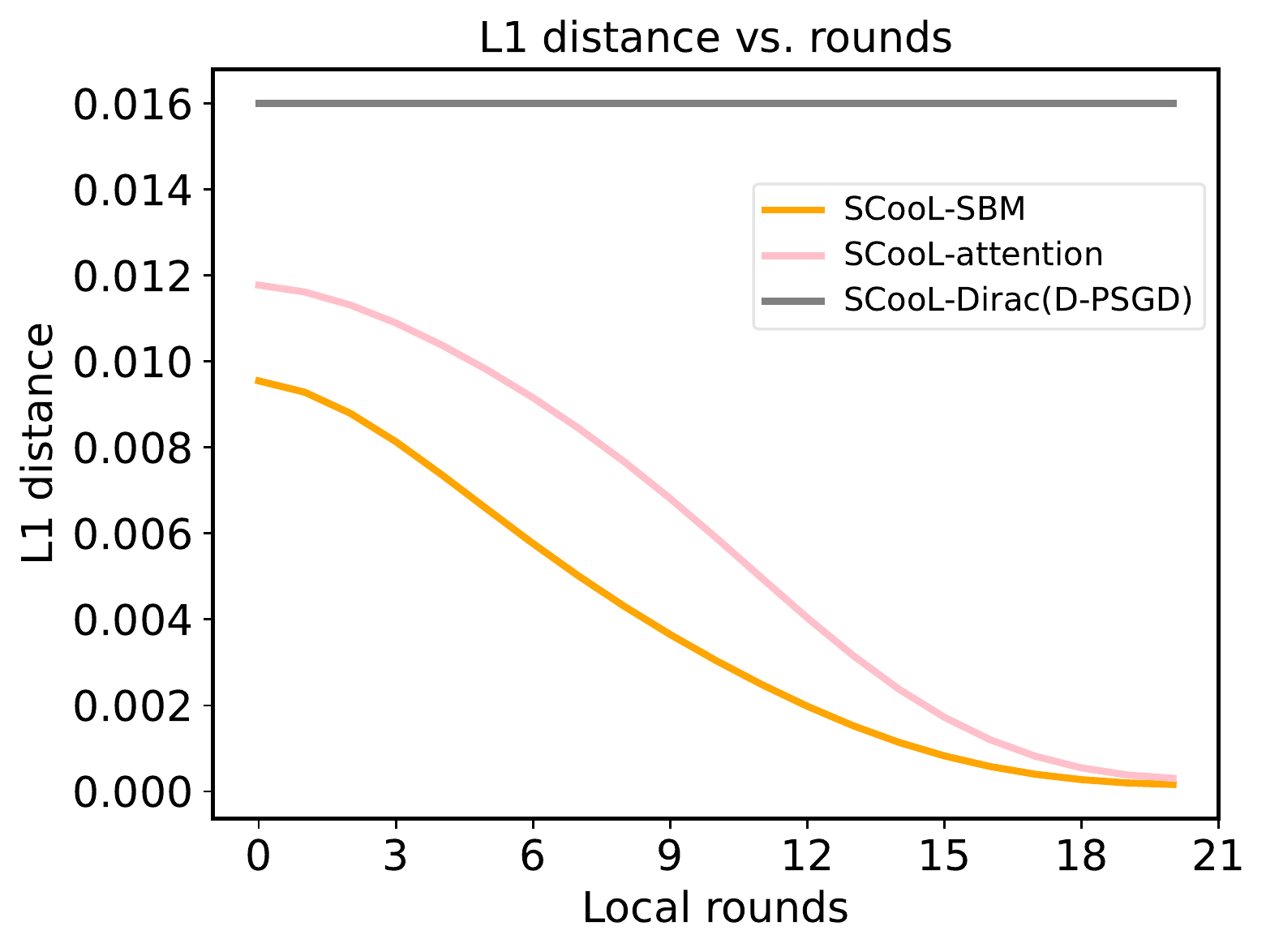}
			\caption{Non-IID SBM setting.}
			\label{fig:SCooL Y on non-iid SBM setting}
		\end{subfigure}
	\begin{subfigure}{0.48\linewidth}
			\includegraphics[width=\linewidth]{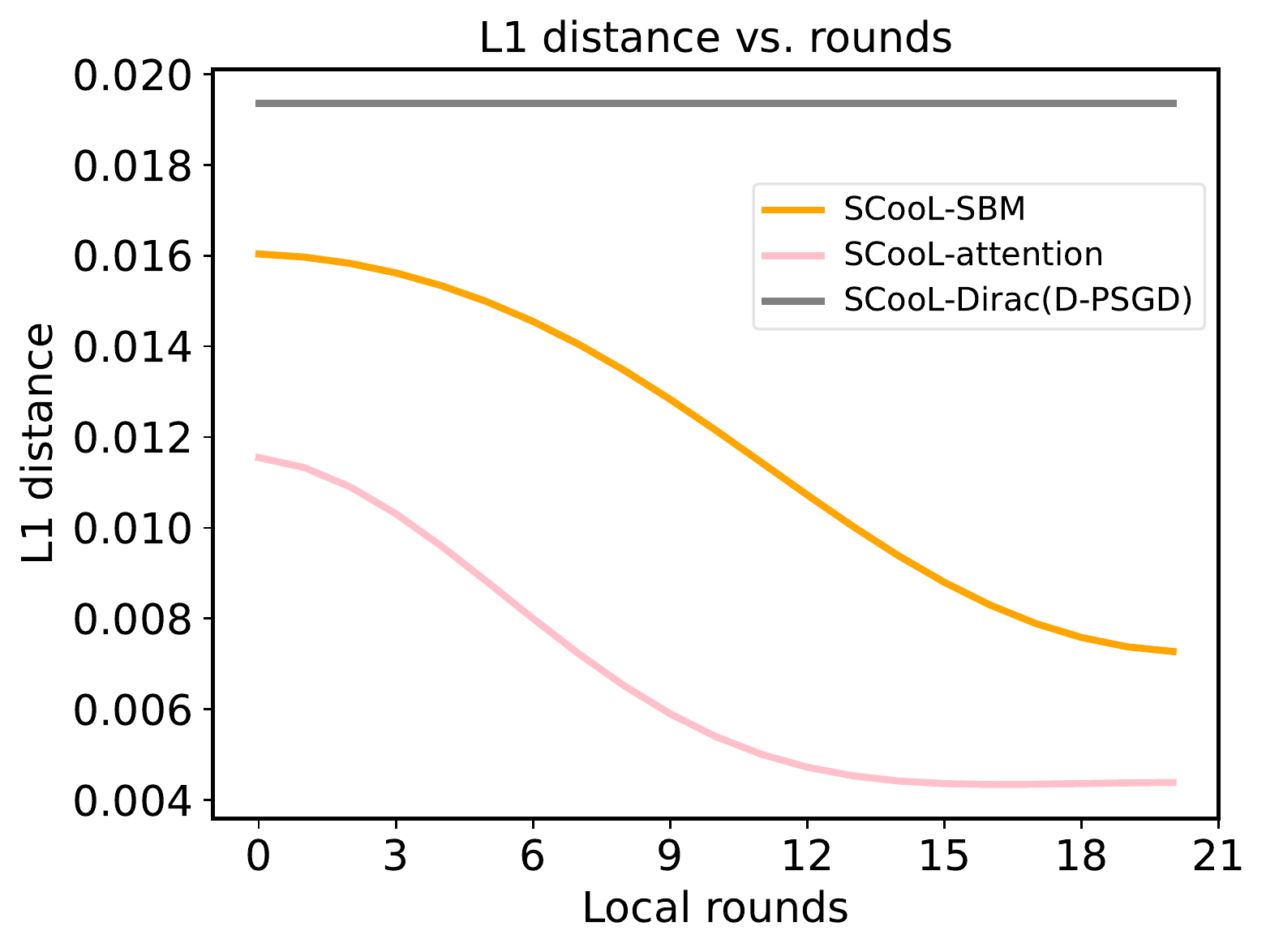}
    			\caption{Non-IID setting.}
			\label{fig:SCooL Y on non-iid general setting}
		\end{subfigure}%
\caption{\footnotesize Mixing weights L1 distance to ground-truth $Y$ in the first 20 training rounds out of totally 100 rounds, (a) for non-IID SBM setting, (b) for non-IID~\cite{mcmahan2017communication} setting. Both SCooL-SBM and SCooL-attention finally converge to generate more precise mixing weights. SCooL-SBM adapts mixing weights faster on non-IID SBM setting, while SCooL-attention adapts faster and converges to lower L1 error in non-IID setting.} 
		\label{fig: Y distance}
	\end{figure}

    In addition, we conduct a quantitative evaluation to the learned cooperation graphs by comparing the mixing weights $w$ with the ground truth $w^*$ used to draw the non-IID tasks. 
    In the ground truth $w^*$, $w^*_{ij} = 1$ for two clients $i$ and $j$ sharing the same data distribution and $w^*_{ij} = 0$ otherwise. 
    We normalize each row in $w^*$ so entries in each row sum up to one. 
    Fig.~\ref{fig:SCooL Y on non-iid SBM setting} shows that the mixing weights in SCooL-SBM converge faster to the ground-truth than SCooL-attention and achieve a similar L1 error because SCooL with SBM prior can better capture SBM-generated graph structures in the non-IID SBM setting. 
    In Fig.~\ref{fig:SCooL Y on non-iid general setting}, SCooL-attention is faster on finding a more accurate cooperation graph due to its attention prior modeling peer-to-peer similarity in the non-IID setting. 

    \paragraph{Communication cost/budget} 
    \begin{figure}[!htbp]
		\begin{center}

   \includegraphics[width=0.7\linewidth]{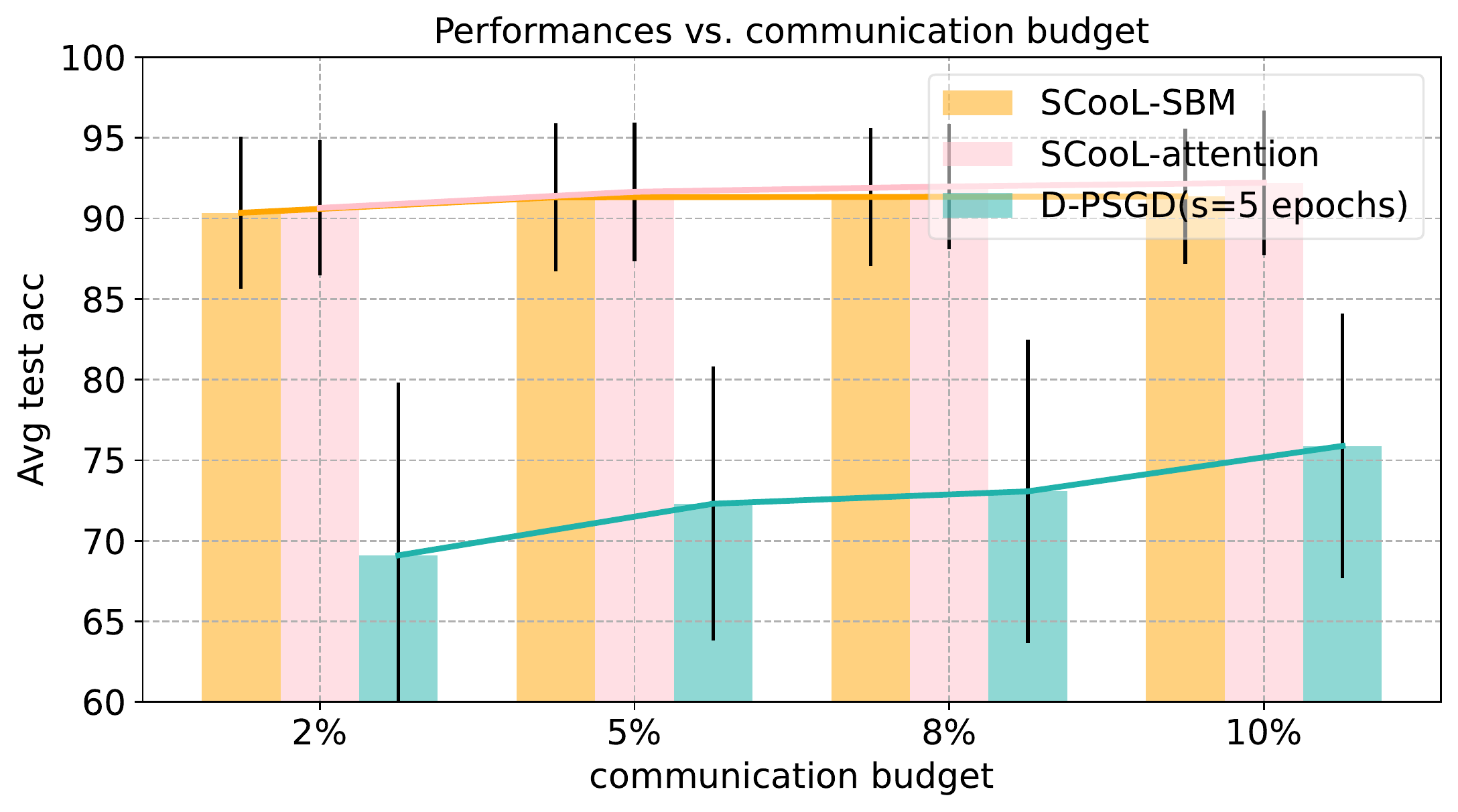}
		\end{center}
		\caption{
			\footnotesize 
			SCooL-SBM and SCooL-attention are robust to communication cost/budget changes. We evaluate each method for every client communicating with 2\%, 5\%, 8\%, or 10\% of other clients.\looseness-1
		}
		\label{fig:communication budget}
	\end{figure}
	The communication cost is often a bottleneck of DL so a sparse topology is usually preferred in practice. In Fig.~\ref{fig:communication budget}, we evaluate the personalization performance of SCooL-SBM, SCooL-attention, and D-PSGD under different communication budgets. Both of our methods achieve almost the same accuracy under different budgets while D-PSGD's performs much poorer and its accuracy highly depends on an increased budget. In contrast, SCooL only requires communication to 2\% neighbours on average to achieve a much higher test accuracy. 

 \section{Conclusion}
We propose a probabilistic modeling scheme``Structured Cooperative Learning (SCooL)'' for decentralized learning of personalized models. SCooL improves the cooperative learning of personalized models across clients by alternately optimizing a cooperation graph $Y$ and the personalized models. We introduce three instantiations of SCooL that adopt different graphical model priors to generate $Y$. They leverage the structural prior among clients to capture an accurate cooperation graph that improves each local model by its neighbors' models. 
We empirically demonstrate the advantages of SCooL over SOTA Federated/Decentralized Learning methods on personalization performance and communication efficiency in different non-IID settings. SCooL is a general framework for efficient knowledge sharing between decentralized agents on a network. It combines the strengths of both the neural networks on local data fitting and graphical models on capturing the cooperation graph structure, leading to interpretable and efficient decentralized learning algorithms with learnable cooperation. In the future work, we are going to study SCooL for partial-model personalization and other non-classification tasks. 

\section*{Acknowledgement}
Shuangtong and Prof. Xinmei Tian are partially supported by NSFC No. 62222117 and the Fundamental Research Funds for the Central Universities under contract WK3490000005. Prof. Dacheng Tao is partially supported by Australian Research Council Project FL-170100117.

\bibliographystyle{unsrtnat}
\bibliography{example_paper}  

\begin{thebibliography}{53}
\providecommand{\natexlab}[1]{#1}
\providecommand{\url}[1]{\texttt{#1}}
\expandafter\ifx\csname urlstyle\endcsname\relax
  \providecommand{\doi}[1]{doi: #1}\else
  \providecommand{\doi}{doi: \begingroup \urlstyle{rm}\Url}\fi

\bibitem[Lian et~al.(2017)Lian, Zhang, Zhang, Hsieh, Zhang, and
  Liu]{NIPS2017_f7552665}
Xiangru Lian, Ce~Zhang, Huan Zhang, Cho-Jui Hsieh, Wei Zhang, and Ji~Liu.
\newblock Can decentralized algorithms outperform centralized algorithms? a
  case study for decentralized parallel stochastic gradient descent.
\newblock In I.~Guyon, U.~V. Luxburg, S.~Bengio, H.~Wallach, R.~Fergus,
  S.~Vishwanathan, and R.~Garnett, editors, \emph{Advances in Neural
  Information Processing Systems}, volume~30. Curran Associates, Inc., 2017.
\newblock URL
  \url{https://proceedings.neurips.cc/paper/2017/file/f75526659f31040afeb61cb7133e4e6d-Paper.pdf}.

\bibitem[McMahan et~al.(2017)McMahan, Moore, Ramage, Hampson, and
  y~Arcas]{mcmahan2017communication}
Brendan McMahan, Eider Moore, Daniel Ramage, Seth Hampson, and Blaise~Aguera
  y~Arcas.
\newblock Communication-efficient learning of deep networks from decentralized
  data.
\newblock In \emph{Artificial intelligence and statistics}, pages 1273--1282.
  PMLR, 2017.

\bibitem[Hsieh et~al.(2020)Hsieh, Phanishayee, Mutlu, and
  Gibbons]{hsieh2020non}
Kevin Hsieh, Amar Phanishayee, Onur Mutlu, and Phillip Gibbons.
\newblock The non-iid data quagmire of decentralized machine learning.
\newblock In \emph{International Conference on Machine Learning}, pages
  4387--4398. PMLR, 2020.

\bibitem[Karimireddy et~al.(2020)Karimireddy, Kale, Mohri, Reddi, Stich, and
  Suresh]{karimireddy2020scaffold}
Sai~Praneeth Karimireddy, Satyen Kale, Mehryar Mohri, Sashank Reddi, Sebastian
  Stich, and Ananda~Theertha Suresh.
\newblock Scaffold: Stochastic controlled averaging for federated learning.
\newblock In \emph{International Conference on Machine Learning}, pages
  5132--5143. PMLR, 2020.

\bibitem[Lin et~al.(2020)Lin, Kong, Stich, and Jaggi]{NEURIPS2020_18df51b9}
Tao Lin, Lingjing Kong, Sebastian~U Stich, and Martin Jaggi.
\newblock Ensemble distillation for robust model fusion in federated learning.
\newblock In H.~Larochelle, M.~Ranzato, R.~Hadsell, M.~F. Balcan, and H.~Lin,
  editors, \emph{Advances in Neural Information Processing Systems}, volume~33,
  pages 2351--2363. Curran Associates, Inc., 2020.
\newblock URL
  \url{https://proceedings.neurips.cc/paper/2020/file/18df51b97ccd68128e994804f3eccc87-Paper.pdf}.

\bibitem[Fraboni et~al.(2021)Fraboni, Vidal, Kameni, and
  Lorenzi]{pmlr-v139-fraboni21a}
Yann Fraboni, Richard Vidal, Laetitia Kameni, and Marco Lorenzi.
\newblock Clustered sampling: Low-variance and improved representativity for
  clients selection in federated learning.
\newblock In Marina Meila and Tong Zhang, editors, \emph{Proceedings of the
  38th International Conference on Machine Learning}, volume 139 of
  \emph{Proceedings of Machine Learning Research}, pages 3407--3416. PMLR,
  18--24 Jul 2021.
\newblock URL \url{https://proceedings.mlr.press/v139/fraboni21a.html}.

\bibitem[Chen and Chao(2021)]{chen2021fedbe}
Hong-You Chen and Wei-Lun Chao.
\newblock Fed{\{}be{\}}: Making bayesian model ensemble applicable to federated
  learning.
\newblock In \emph{International Conference on Learning Representations}, 2021.
\newblock URL \url{https://openreview.net/forum?id=dgtpE6gKjHn}.

\bibitem[Wang et~al.(2020)Wang, Yurochkin, Sun, Papailiopoulos, and
  Khazaeni]{Wang2020Federated}
Hongyi Wang, Mikhail Yurochkin, Yuekai Sun, Dimitris Papailiopoulos, and
  Yasaman Khazaeni.
\newblock Federated learning with matched averaging.
\newblock In \emph{International Conference on Learning Representations}, 2020.
\newblock URL \url{https://openreview.net/forum?id=BkluqlSFDS}.

\bibitem[Balakrishnan et~al.(2022)Balakrishnan, Li, Zhou, Himayat, Smith, and
  Bilmes]{balakrishnan2022diverse}
Ravikumar Balakrishnan, Tian Li, Tianyi Zhou, Nageen Himayat, Virginia Smith,
  and Jeff Bilmes.
\newblock Diverse client selection for federated learning via submodular
  maximization.
\newblock In \emph{International Conference on Learning Representations}, 2022.
\newblock URL \url{https://openreview.net/forum?id=nwKXyFvaUm}.

\bibitem[Acar et~al.(2021)Acar, Zhao, Matas, Mattina, Whatmough, and
  Saligrama]{acar2021federated}
Durmus Alp~Emre Acar, Yue Zhao, Ramon Matas, Matthew Mattina, Paul Whatmough,
  and Venkatesh Saligrama.
\newblock Federated learning based on dynamic regularization.
\newblock In \emph{International Conference on Learning Representations}, 2021.
\newblock URL \url{https://openreview.net/forum?id=B7v4QMR6Z9w}.

\bibitem[Li et~al.(2018)Li, Sahu, Zaheer, Sanjabi, Talwalkar, and
  Smith]{li2018federated}
Tian Li, Anit~Kumar Sahu, Manzil Zaheer, Maziar Sanjabi, Ameet Talwalkar, and
  Virginia Smith.
\newblock Federated optimization in heterogeneous networks.
\newblock \emph{arXiv preprint arXiv:1812.06127}, 2018.

\bibitem[Xu et~al.(2022)Xu, Hong, Huang, and Jiang]{xu2022acceleration}
Chencheng Xu, Zhiwei Hong, Minlie Huang, and Tao Jiang.
\newblock Acceleration of federated learning with alleviated forgetting in
  local training.
\newblock In \emph{International Conference on Learning Representations}, 2022.
\newblock URL \url{https://openreview.net/forum?id=541PxiEKN3F}.

\bibitem[Li et~al.(2021{\natexlab{a}})Li, Hu, Beirami, and Smith]{li2021ditto}
Tian Li, Shengyuan Hu, Ahmad Beirami, and Virginia Smith.
\newblock Ditto: Fair and robust federated learning through personalization.
\newblock In \emph{International Conference on Machine Learning}, pages
  6357--6368. PMLR, 2021{\natexlab{a}}.

\bibitem[T.~Dinh et~al.(2020)T.~Dinh, Tran, and Nguyen]{NEURIPS2020_f4f1f13c}
Canh T.~Dinh, Nguyen Tran, and Josh Nguyen.
\newblock Personalized federated learning with moreau envelopes.
\newblock In H.~Larochelle, M.~Ranzato, R.~Hadsell, M.~F. Balcan, and H.~Lin,
  editors, \emph{Advances in Neural Information Processing Systems}, volume~33,
  pages 21394--21405. Curran Associates, Inc., 2020.
\newblock URL
  \url{https://proceedings.neurips.cc/paper/2020/file/f4f1f13c8289ac1b1ee0ff176b56fc60-Paper.pdf}.

\bibitem[Sattler et~al.(2020)Sattler, M{\"u}ller, and
  Samek]{sattler2020clustered}
Felix Sattler, Klaus-Robert M{\"u}ller, and Wojciech Samek.
\newblock Clustered federated learning: Model-agnostic distributed multitask
  optimization under privacy constraints.
\newblock \emph{IEEE transactions on neural networks and learning systems},
  2020.

\bibitem[Ghosh et~al.(2020)Ghosh, Chung, Yin, and
  Ramchandran]{ghosh2020efficient}
Avishek Ghosh, Jichan Chung, Dong Yin, and Kannan Ramchandran.
\newblock An efficient framework for clustered federated learning.
\newblock \emph{arXiv preprint arXiv:2006.04088}, 2020.

\bibitem[Xie et~al.(2021)Xie, Long, Shen, Zhou, Wang, Jiang, and
  Zhang]{xie2021multi}
Ming Xie, Guodong Long, Tao Shen, Tianyi Zhou, Xianzhi Wang, Jing Jiang, and
  Chengqi Zhang.
\newblock Multi-center federated learning.
\newblock \emph{arXiv preprint arXiv:2108.08647}, 2021.

\bibitem[Long et~al.(2022)Long, Xie, Shen, Zhou, Wang, and
  Jiang]{Multi-Center_Federated_Learning}
Guodong Long, Ming Xie, Tao Shen, Tianyi Zhou, Xianzhi Wang, and Jing Jiang.
\newblock Multi-center federated learning: clients clustering for better
  personalization.
\newblock \emph{World Wide Web Journal (Springer)}, 2022.

\bibitem[Li et~al.(2021{\natexlab{b}})Li, JIANG, Zhang, Kamp, and
  Dou]{li2021fedbn}
Xiaoxiao Li, Meirui JIANG, Xiaofei Zhang, Michael Kamp, and Qi~Dou.
\newblock Fed{BN}: Federated learning on non-{IID} features via local batch
  normalization.
\newblock In \emph{International Conference on Learning Representations},
  2021{\natexlab{b}}.
\newblock URL \url{https://openreview.net/forum?id=6YEQUn0QICG}.

\bibitem[Liang et~al.(2020)Liang, Liu, Ziyin, Allen, Auerbach, Brent,
  Salakhutdinov, and Morency]{liang2020think}
Paul~Pu Liang, Terrance Liu, Liu Ziyin, Nicholas~B Allen, Randy~P Auerbach,
  David Brent, Ruslan Salakhutdinov, and Louis-Philippe Morency.
\newblock Think locally, act globally: Federated learning with local and global
  representations.
\newblock \emph{arXiv preprint arXiv:2001.01523}, 2020.

\bibitem[Collins et~al.(2021)Collins, Hassani, Mokhtari, and
  Shakkottai]{pmlr-v139-collins21a}
Liam Collins, Hamed Hassani, Aryan Mokhtari, and Sanjay Shakkottai.
\newblock Exploiting shared representations for personalized federated
  learning.
\newblock In Marina Meila and Tong Zhang, editors, \emph{Proceedings of the
  38th International Conference on Machine Learning}, volume 139 of
  \emph{Proceedings of Machine Learning Research}, pages 2089--2099. PMLR,
  18--24 Jul 2021.
\newblock URL \url{https://proceedings.mlr.press/v139/collins21a.html}.

\bibitem[Oh et~al.(2022)Oh, Kim, and Yun]{oh2022fedbabu}
Jaehoon Oh, SangMook Kim, and Se-Young Yun.
\newblock Fed{BABU}: Toward enhanced representation for federated image
  classification.
\newblock In \emph{International Conference on Learning Representations}, 2022.
\newblock URL \url{https://openreview.net/forum?id=HuaYQfggn5u}.

\bibitem[Zhang et~al.(2023)Zhang, Long, Zhou, Yan, Zhang, Zhang, and
  Yang]{DualPersonalization}
Chunxu Zhang, Guodong Long, Tianyi Zhou, Peng Yan, Zijian Zhang, Chengqi Zhang,
  and Bo~Yang.
\newblock Dual personalization on federated recommendation.
\newblock In \emph{International Joint Conference on Artificial Intelligence
  (IJCAI)}, 2023.

\bibitem[Zhu et~al.(2021)Zhu, Hong, and Zhou]{zhu2021data}
Zhuangdi Zhu, Junyuan Hong, and Jiayu Zhou.
\newblock Data-free knowledge distillation for heterogeneous federated
  learning.
\newblock In Marina Meila and Tong Zhang, editors, \emph{Proceedings of the
  38th International Conference on Machine Learning}, volume 139 of
  \emph{Proceedings of Machine Learning Research}, pages 12878--12889. PMLR,
  18--24 Jul 2021.
\newblock URL \url{https://proceedings.mlr.press/v139/zhu21b.html}.

\bibitem[Afonin and Karimireddy(2022)]{afonin2022towards}
Andrei Afonin and Sai~Praneeth Karimireddy.
\newblock Towards model agnostic federated learning using knowledge
  distillation.
\newblock In \emph{International Conference on Learning Representations}, 2022.
\newblock URL \url{https://openreview.net/forum?id=lQI_mZjvBxj}.

\bibitem[Fallah et~al.(2020)Fallah, Mokhtari, and
  Ozdaglar]{fallah2020personalized}
Alireza Fallah, Aryan Mokhtari, and Asuman Ozdaglar.
\newblock Personalized federated learning with theoretical guarantees: A
  model-agnostic meta-learning approach.
\newblock \emph{Advances in Neural Information Processing Systems}, 33, 2020.

\bibitem[Shamsian et~al.(2021)Shamsian, Navon, Fetaya, and
  Chechik]{pmlr-v139-shamsian21a}
Aviv Shamsian, Aviv Navon, Ethan Fetaya, and Gal Chechik.
\newblock Personalized federated learning using hypernetworks.
\newblock In Marina Meila and Tong Zhang, editors, \emph{Proceedings of the
  38th International Conference on Machine Learning}, volume 139 of
  \emph{Proceedings of Machine Learning Research}, pages 9489--9502. PMLR,
  18--24 Jul 2021.
\newblock URL \url{https://proceedings.mlr.press/v139/shamsian21a.html}.

\bibitem[Tan et~al.(2022{\natexlab{a}})Tan, Long, Liu, Zhou, Lu, Jiang, and
  Zhang]{tan2022fedproto}
Yue Tan, Guodong Long, Lu~Liu, Tianyi Zhou, Qinghua Lu, Jing Jiang, and Chengqi
  Zhang.
\newblock Fedproto: Federated prototype learning across heterogeneous clients.
\newblock In \emph{AAAI Conference on Artificial Intelligence}, volume~1,
  2022{\natexlab{a}}.

\bibitem[Tan et~al.(2022{\natexlab{b}})Tan, Long, Ma, Liu, Zhou, and
  Jiang]{tan2022federated}
Yue Tan, Guodong Long, Jie Ma, Lu~Liu, Tianyi Zhou, and Jing Jiang.
\newblock Federated learning from pre-trained models: A contrastive learning
  approach.
\newblock In Alice~H. Oh, Alekh Agarwal, Danielle Belgrave, and Kyunghyun Cho,
  editors, \emph{Advances in Neural Information Processing Systems},
  2022{\natexlab{b}}.
\newblock URL \url{https://openreview.net/forum?id=mhQLcMjWw75}.

\bibitem[Dai et~al.(2022)Dai, Shen, He, Tian, and
  Tao]{DBLP:conf/icml/Dai0H0T22}
Rong Dai, Li~Shen, Fengxiang He, Xinmei Tian, and Dacheng Tao.
\newblock Dispfl: Towards communication-efficient personalized federated
  learning via decentralized sparse training.
\newblock In Kamalika Chaudhuri, Stefanie Jegelka, Le~Song, Csaba
  Szepesv{\'{a}}ri, Gang Niu, and Sivan Sabato, editors, \emph{International
  Conference on Machine Learning, {ICML} 2022, 17-23 July 2022, Baltimore,
  Maryland, {USA}}, volume 162 of \emph{Proceedings of Machine Learning
  Research}, pages 4587--4604. {PMLR}, 2022.
\newblock URL \url{https://proceedings.mlr.press/v162/dai22b.html}.

\bibitem[Chen et~al.(2022)Chen, Long, Wu, Zhou, and
  Jiang]{Personalized_Federated_Learning_With_Structural_Information}
Fengwen Chen, Guodong Long, Zonghan Wu, Tianyi Zhou, and Jing Jiang.
\newblock Personalized federated learning with structural information.
\newblock In \emph{International Joint Conference on Artificial Intelligence
  (IJCAI)}, 2022.

\bibitem[Blot et~al.(2016)Blot, Picard, Cord, and Thome]{blot2016gossip}
Michael Blot, David Picard, Matthieu Cord, and Nicolas Thome.
\newblock Gossip training for deep learning.
\newblock \emph{arXiv preprint arXiv:1611.09726}, 2016.

\bibitem[Jiang et~al.(2017)Jiang, Balu, Hegde, and Sarkar]{NIPS2017_a74c3bae}
Zhanhong Jiang, Aditya Balu, Chinmay Hegde, and Soumik Sarkar.
\newblock Collaborative deep learning in fixed topology networks.
\newblock In I.~Guyon, U.~V. Luxburg, S.~Bengio, H.~Wallach, R.~Fergus,
  S.~Vishwanathan, and R.~Garnett, editors, \emph{Advances in Neural
  Information Processing Systems}, volume~30. Curran Associates, Inc., 2017.
\newblock URL
  \url{https://proceedings.neurips.cc/paper/2017/file/a74c3bae3e13616104c1b25f9da1f11f-Paper.pdf}.

\bibitem[Lin et~al.(2021)Lin, Karimireddy, Stich, and Jaggi]{lin2021quasi}
Tao Lin, Sai~Praneeth Karimireddy, Sebastian Stich, and Martin Jaggi.
\newblock Quasi-global momentum: Accelerating decentralized deep learning on
  heterogeneous data.
\newblock In Marina Meila and Tong Zhang, editors, \emph{Proceedings of the
  38th International Conference on Machine Learning}, volume 139 of
  \emph{Proceedings of Machine Learning Research}, pages 6654--6665. PMLR,
  18--24 Jul 2021.
\newblock URL \url{https://proceedings.mlr.press/v139/lin21c.html}.

\bibitem[Khawatmi et~al.(2017)Khawatmi, Sayed, and
  Zoubir]{khawatmi2017decentralized}
Sahar Khawatmi, Ali~H Sayed, and Abdelhak~M Zoubir.
\newblock Decentralized clustering and linking by networked agents.
\newblock \emph{IEEE Transactions on Signal Processing}, 65\penalty0
  (13):\penalty0 3526--3537, 2017.

\bibitem[Esfandiari et~al.(2021)Esfandiari, Tan, Jiang, Balu, Herron, Hegde,
  and Sarkar]{esfandiari2021cross}
Yasaman Esfandiari, Sin~Yong Tan, Zhanhong Jiang, Aditya Balu, Ethan Herron,
  Chinmay Hegde, and Soumik Sarkar.
\newblock Cross-gradient aggregation for decentralized learning from non-iid
  data.
\newblock In Marina Meila and Tong Zhang, editors, \emph{Proceedings of the
  38th International Conference on Machine Learning}, volume 139 of
  \emph{Proceedings of Machine Learning Research}, pages 3036--3046. PMLR,
  18--24 Jul 2021.
\newblock URL \url{https://proceedings.mlr.press/v139/esfandiari21a.html}.

\bibitem[Huang et~al.(2022)Huang, Sun, Zhu, Yan, and
  Xu]{DBLP:conf/icml/HuangSZYX22}
Yan Huang, Ying Sun, Zehan Zhu, Changzhi Yan, and Jinming Xu.
\newblock Tackling data heterogeneity: {A} new unified framework for
  decentralized {SGD} with sample-induced topology.
\newblock In Kamalika Chaudhuri, Stefanie Jegelka, Le~Song, Csaba
  Szepesv{\'{a}}ri, Gang Niu, and Sivan Sabato, editors, \emph{International
  Conference on Machine Learning, {ICML} 2022, 17-23 July 2022, Baltimore,
  Maryland, {USA}}, volume 162 of \emph{Proceedings of Machine Learning
  Research}, pages 9310--9345. {PMLR}, 2022.
\newblock URL \url{https://proceedings.mlr.press/v162/huang22i.html}.

\bibitem[Yuan et~al.(2022)Yuan, Huang, Chen, Zhang, Zhang, and
  Pan]{yuan2022revisiting}
Kun Yuan, Xinmeng Huang, Yiming Chen, Xiaohan Zhang, Yingya Zhang, and Pan Pan.
\newblock Revisiting optimal convergence rate for smooth and non-convex
  stochastic decentralized optimization.
\newblock In Alice~H. Oh, Alekh Agarwal, Danielle Belgrave, and Kyunghyun Cho,
  editors, \emph{Advances in Neural Information Processing Systems}, 2022.
\newblock URL \url{https://openreview.net/forum?id=eHePKMLuNmy}.

\bibitem[Song et~al.(2022)Song, Li, Jin, Shi, Yan, Yin, and
  Yuan]{song2022communicationefficient}
Zhuoqing Song, Weijian Li, Kexin Jin, Lei Shi, Ming Yan, Wotao Yin, and Kun
  Yuan.
\newblock Communication-efficient topologies for decentralized learning with
  \$o(1)\$ consensus rate.
\newblock In Alice~H. Oh, Alekh Agarwal, Danielle Belgrave, and Kyunghyun Cho,
  editors, \emph{Advances in Neural Information Processing Systems}, 2022.
\newblock URL \url{https://openreview.net/forum?id=AyiiHcRzTd}.

\bibitem[Vogels et~al.(2022)Vogels, Hendrikx, and Jaggi]{vogels2022beyond}
Thijs Vogels, Hadrien Hendrikx, and Martin Jaggi.
\newblock Beyond spectral gap: the role of the topology in decentralized
  learning.
\newblock In Alice~H. Oh, Alekh Agarwal, Danielle Belgrave, and Kyunghyun Cho,
  editors, \emph{Advances in Neural Information Processing Systems}, 2022.
\newblock URL \url{https://openreview.net/forum?id=AQgmyyEWg8}.

\bibitem[Lu et~al.(2022)Lu, Cui, Squillante, Kingsbury, and
  Horesh]{lu2022decentralized}
Songtao Lu, Xiaodong Cui, Mark~S Squillante, Brian Kingsbury, and Lior Horesh.
\newblock Decentralized bilevel optimization for personalized client learning.
\newblock In \emph{ICASSP 2022-2022 IEEE International Conference on Acoustics,
  Speech and Signal Processing (ICASSP)}, pages 5543--5547. IEEE, 2022.

\bibitem[Jordan et~al.(1999)Jordan, Ghahramani, Jaakkola, and
  Saul]{DBLP:journals/ml/JordanGJS99}
Michael~I. Jordan, Zoubin Ghahramani, Tommi~S. Jaakkola, and Lawrence~K. Saul.
\newblock An introduction to variational methods for graphical models.
\newblock \emph{Mach. Learn.}, 37\penalty0 (2):\penalty0 183--233, 1999.
\newblock \doi{10.1023/A:1007665907178}.
\newblock URL \url{https://doi.org/10.1023/A:1007665907178}.

\bibitem[Belkin et~al.(2006)Belkin, Niyogi, and Sindhwani]{belkin2006manifold}
Mikhail Belkin, Partha Niyogi, and Vikas Sindhwani.
\newblock Manifold regularization: A geometric framework for learning from
  labeled and unlabeled examples.
\newblock \emph{Journal of machine learning research}, 7\penalty0 (11), 2006.

\bibitem[Holland et~al.(1983)Holland, Laskey, and
  Leinhardt]{holland1983stochastic}
Paul~W Holland, Kathryn~Blackmond Laskey, and Samuel Leinhardt.
\newblock Stochastic blockmodels: First steps.
\newblock \emph{Social networks}, 5\penalty0 (2):\penalty0 109--137, 1983.

\bibitem[Krizhevsky et~al.(2009)Krizhevsky, Hinton,
  et~al.]{krizhevsky2009learning}
Alex Krizhevsky, Geoffrey Hinton, et~al.
\newblock Learning multiple layers of features from tiny images.
\newblock 2009.

\bibitem[Ravi and Larochelle(2017)]{ravi2017optimization}
Sachin Ravi and Hugo Larochelle.
\newblock Optimization as a model for few-shot learning.
\newblock In \emph{International Conference on Learning Representations}, 2017.

\bibitem[Zhang et~al.(2021)Zhang, Sapra, Fidler, Yeung, and
  Alvarez]{zhang2021personalized}
Michael Zhang, Karan Sapra, Sanja Fidler, Serena Yeung, and Jose~M. Alvarez.
\newblock Personalized federated learning with first order model optimization.
\newblock In \emph{International Conference on Learning Representations}, 2021.
\newblock URL \url{https://openreview.net/forum?id=ehJqJQk9cw}.

\bibitem[Finn et~al.(2017)Finn, Abbeel, and Levine]{finn2017model}
Chelsea Finn, Pieter Abbeel, and Sergey Levine.
\newblock Model-agnostic meta-learning for fast adaptation of deep networks.
\newblock In \emph{International Conference on Machine Learning}, pages
  1126--1135. PMLR, 2017.

\bibitem[Ioffe and Szegedy(2015)]{pmlr-v37-ioffe15}
Sergey Ioffe and Christian Szegedy.
\newblock Batch normalization: Accelerating deep network training by reducing
  internal covariate shift.
\newblock In Francis Bach and David Blei, editors, \emph{Proceedings of the
  32nd International Conference on Machine Learning}, volume~37 of
  \emph{Proceedings of Machine Learning Research}, pages 448--456, Lille,
  France, 07--09 Jul 2015. PMLR.
\newblock URL \url{https://proceedings.mlr.press/v37/ioffe15.html}.

\bibitem[Wu and He(2018)]{Wu_2018_ECCV}
Yuxin Wu and Kaiming He.
\newblock Group normalization.
\newblock In \emph{Proceedings of the European Conference on Computer Vision
  (ECCV)}, September 2018.

\bibitem[Li et~al.(2022)Li, Zhou, Tian, and Tao]{li2022learning}
Shuangtong Li, Tianyi Zhou, Xinmei Tian, and Dacheng Tao.
\newblock Learning to collaborate in decentralized learning of personalized
  models.
\newblock In \emph{IEEE/CVF Conference on Computer Vision and Pattern
  Recognition (CVPR)}, 2022.

\bibitem[Zantedeschi et~al.(2020)Zantedeschi, Bellet, and
  Tommasi]{zantedeschi2020fully}
Valentina Zantedeschi, Aur{\'e}lien Bellet, and Marc Tommasi.
\newblock Fully decentralized joint learning of personalized models and
  collaboration graphs.
\newblock In \emph{International Conference on Artificial Intelligence and
  Statistics}, pages 864--874. PMLR, 2020.

\bibitem[Kingma and Ba(2014)]{kingma2014adam}
Diederik~P Kingma and Jimmy Ba.
\newblock Adam: A method for stochastic optimization.
\newblock \emph{arXiv preprint arXiv:1412.6980}, 2014.

\end{thebibliography}






	\newpage
	\appendix
	\onecolumn
 \section{Notations}
  \begin{table}[htbp]
		\caption{Notations used in this paper.}
		\label{table:noniid}
		\begin{center}
			\begin{tabular}{cc}
			\toprule
				\bf Notation & \bf Description
				\\ \midrule
				$\theta_i$ & Personalized model on the i'th client\\
    $D_i$ & Dataset for the i'th client \\
    $Y$ & cooperation graph \\
    $\Phi$ & Observable variables set \\
    $Z$ & Latent variables set \\
    $\beta$ & Variational parameters of latent variables $Z$ \\
    $w$ & Variational parameter of cooperation graph $Y$ \\
    $\pi_i$ & Prior membership distribution of client-$i$ in SBM model \\
    $\alpha_i$ & Parameter for Dirichlet distribution of $\pi_i$\\ 
    $z_i$ & Membership indicator for client-$i$ in SBM model\\
    $B$ & Pairwise correlations of memberships in SBM model\\
    $\gamma$ & Variational parameter of $\pi$\\
    $\Omega$ & Variational parameter of $z$ \\
    $\phi$ & Learnable neural network parameters for attention prior \\

				\bottomrule			
			\end{tabular}
		\end{center}
 \end{table}

	\section{D-PSGD algorithm}
	\label{sec:D-PSGD}
	
	\begin{algorithm}[h]
		\caption{D-PSGD}
		\label{alg:D-PSGD}
		\begin{algorithmic}
			\STATE {\bfseries Require:} initial point $x_{0, i} = x_0$, step length $\gamma$, weight matrix $w$, and number of iterations $T$
			\FOR{$t = 0$ {\bfseries to} $T - 1$}
			\STATE Randomly sample $\zeta_{t, i}$ from local data of the i-th client
			\STATE Compute a local stochastic gradient based on $\zeta_{t, i}$ and current optimization variable $x_{t, i}:\nabla F_i(x_{t,i};\zeta_{t,i})$
			\STATE Compute the neighborhood weighted average by fetching optimization variables from neighbors: $x_{t + \frac{1}{2}, i} = \sum_{j = 1}^K w_{ij}x_{t, j}$
			\STATE Update the local optimization variable $x_{t+1, i}\leftarrow x_{t+\frac{1}{2}, i} - \gamma \nabla F_i(x_{t,i};\zeta_{t,i})$
			\ENDFOR
			\STATE {\bfseries Output:} $\frac{1}{K}\sum_{i = 1}^K x_{T,i}$
		\end{algorithmic}
	\end{algorithm}

\section{Experimental Details}
\subsection{Implementation Details of SCooL-SBM and SCooL-attention}\label{sec:SCooL-implementation}
	We apply a lightweight fully connected network of two layers with output dimensions $(10,5)$ as the encoder network $\phi$ in SCooL-attention. 
	We use Adam~\cite{kingma2014adam} with learning rate of 0.1 and weight decay of 0.01 to train both SCooL-SBM and SCooL-attention. We apply Algorithm~\ref{alg:SCooL algorithm} and train all the local models for T=100 rounds with 5 epochs of local SGD per round. To achieve a sparse topology, we sort the learned $w_{i*}$ for each client, and remove $90\%$ neighbors with the smallest $w_{ij}$ for each client after $10$ rounds. 

The update rule of (\ref{equ:SBM theta update}) and (\ref{equ:attention theta update}) requires local model $\theta_i$ to calculate gradients on other clients' dataset $\theta_j$. When local data is not allowed to share across clients, we can follow a ``cross-gradient'' fashion: sending model $\theta_i$ to client-$j$, who then computes gradient on its own data and then sends $\nabla L(D_j;\theta_i)$ back to client-$i$. However, this method requires twice the communication cost of classical decentralized learning algorithms such as D-PSGD. To avoid such cross gradient terms, we can approximate $\nabla L(D_j;\theta_i)$ using $\nabla L(D_j;\theta_j)$ according to first-order Taylor expansion:
\begin{equation}
    L(\theta_i;D_j) = L(\theta_j;D_j) + \nabla L(\theta_j;D_j) (\theta_i - \theta_j) + O(||\theta_i - \theta_j||^2)
\end{equation}
When $\theta_i$ and $\theta_j$ are close to each other, i.e., $O(||\theta_i - \theta_j||^2)$ is small, we can ignore the second-order term and get an approximation with small approximation error, i.e.,
\begin{equation}
    \nabla_{\theta_i} L(\theta_i;D_j) \approx \nabla _{\theta_j}L(\theta_j;D_j).
\end{equation}
In practice, if we initialize all personalized models from the same point, clients with similar or the same tasks tend to have similar optimization paths and thus their models' distance can be upper bounded. For clients with distinct tasks, the learned $w_{ij}$ tends to be close to zero so the approximation error does not result in a notable difference in the model updates in M-step. In experiments, we find that the mixing weights converge quickly and precisely capture task relationships among clients. As shown in Table~\ref{tab:main result}, this approximation already achieves promising personalization performances.

In practice, the training loss of $\log P(D_j|\theta_i)$ in equation (\ref{equ:SBM w update})(\ref{equ:attention w update}) can vary in magnitude  in different training stages, causing the learned $w$ "over-smooth" or "over-sharp" in certain epochs due to the nature of Sigmoid/ Sofrmax. To tackle this issue, we use Sigmoid/ Softmax with temperature factor. The temperature factor is kept fixed during the whole training phase and selected as a hyperparameter.  

\subsection{Details of Decentralized Learning Baselines}\label{sec: decentralized details}
We train our SCooL models with a communication period of 5 epochs. 
Since the DL baselines, i.e. D-PSGD, CGA, SPDB, are originally proposed to only run one local SGD step per round on a single mini-batch, we evaluate them with two settings, i.e., one local step per round and $5$ local epochs per round, and we apply more rounds for the former to match the total local epochs (i.e., 500 epochs) of other methods. To match the communication cost of our methods, we extend the ring and bipartite topology used in previous DL works~\cite{esfandiari2021cross} to increase the number of neighbors for each client. Specifically, we study (1) a ``group-ring'' topology that connects two clients $i$ and $j$ if $|i-j|\leq \frac{(K-K_0)}{2}$ or $K-|i-j|\leq \frac{(K-K_0)}{2}$; and (2) a generalized bipartite topology that randomly partitions all clients into two groups and then connect each client in a group to $10$ clients randomly drawn from the other group. In our experiments, they both outperform their original versions with fewer neighbors and communications. Hence, in the following, we always report the best result among all the four types of topology for each DL baseline.  
	

	\section{SCooL-SBM derivation}\label{sec:SCooL-SBM-EM}
	\begin{equation}
		\begin{aligned}
			P(\theta_{1:K}|D_{1:K}) \propto P(\theta_{1:K},D_{1:K}) \\= \int P(D_{1:K}|\theta_{1:K},Y)P(\theta_{1:K}|Y)P(Y)dY
		\end{aligned}
\end{equation}
	with
	\begin{equation}
		\begin{aligned}
			&P(\theta_{1:K},Y)=P(\theta_{1:K})P(Y)\\
			&P(Y) =\int P(Y,\vec{\pi}_{1:K},\vec{Z}_{1:K}|\vec{\alpha},B)d(\vec{\pi}_{1:K},\vec{Z}_{1:K}) \\
			&P(\theta_{1:K})\propto \exp(-\frac{\lambda}{2}\sum_{i}||\theta_i||^2)\\
			&P(D_{1:K}|\theta_{1:K},Y)=
			\prod_{i=1}^K P(D_{1:K}|\theta_i,Y)=
			\prod_{i=1}^k{\bigg(P(D_i|\theta_i)\prod_{j\neq i ,Y_{ij}=1}}{P(D_j|\theta_i)}\bigg)
		\end{aligned}
	\end{equation}
	\subsubsection{Objective}
	Modelling $P(Y)$ as SBM:
	\begin{itemize}
		\item For each client $i \in [K]$:
		\begin{itemize}
			\item Draw a $M$ dimensional membership vector $\vec{\pi}_i \sim
			\textrm{Dirichlet}( \vec\alpha )$.
			\item Draw membership indicator
			$\vec z_{i} ~ \sim {\rm Multinomial}(\vec\pi_{i})$.
		\end{itemize}
		\item For each pair of clients $(i,j) \in K \times K$:
		\begin{itemize}
			\item Sample the value of their interaction, $Y_{ij} \sim {\rm
				Bernoulli}(\vec z_{i}^{\ T} B ~ \vec z_{j})$.
		\end{itemize}
	\end{itemize}
	
	Under the SBM, the marginal distribution of $Y$ is:
	\begin{equation}
		\begin{aligned}
			&P(Y|\vec{\alpha},B)=\int P(Y,\vec{\pi}_{1:K},\vec{Z}_{1:K}|\vec{\alpha},B)d(\vec{\pi}_{1:K},\vec{Z}_{1:K})\\
			&= \int \bigg(\prod_{ij}{P(Y_{ij}|\vec{z}_{i},\vec{z}_{j},B)} \prod_i{P(\vec{\pi}_i|\vec{\alpha}_i)P(\vec{z}_i|\vec{\pi}_i)}\bigg)d(\vec{\pi}_{1:K},\vec{Z}_{1:K})
		\end{aligned}
	\end{equation}
	
	Our final objective is:
	\begin{equation}
		\begin{aligned}
			&P(\theta_{1:K}|D_{1:K}) \propto P(\theta_{1:K},D_{1:K}) \\
			= &\int P(D_{1:K}|\theta_{1:K},Y)P(\theta_{1:K})P(Y)dY \\
			= &\int
			\prod_{i=1}^k{\bigg(P(D_i|\theta_i)\prod_{j\neq i ,Y_{ij}=1}}{P(D_j|\theta_i)}\bigg) \exp(-\frac{\lambda}{2}\sum_{i}||\theta_i||^2)\\
			&\bigg(\prod_{ij}{P(Y_{ij}|\vec{z}_{i},\vec{z}_{j},B)} \prod_i{P(\vec{\pi}_i|\vec{\alpha}_i)P(\vec{z}_i|\vec{\pi}_i)}\bigg)d(\vec{\pi}_{1:K},\vec{Z}_{1:K}, Y)
		\end{aligned}
	\end{equation}
	\subsubsection{Optimization}
	\paragraph{ELBO}
	Rewrite $R_1 = (\vec{\pi}_{1:K},\vec{Z}_{1:K}, Y)$, $R_2 = (D_{1:K})$, $R_3=(\theta_{1:K},\vec{\alpha},B)$.
	\begin{equation}
		\begin{aligned}
			\log P(R_2,R_3)&
			=\log \int P(R_1, R_2,R_3)d R_1 \\
			&=\log \int q(R_1)\frac{P(R_1, R_2,R_3)}{q(R_1)}dR_1 \\
			&\geq \int q(R_1) \log \frac{P(R_1, R_2,R_3)}{q(R_1)}dR_1 \\
			&=\mathbb{E}_q \log P(R_1, R_2,R_3) - \log q(R_1) =: \mathbb{L}(q,R_3)
		\end{aligned}
	\end{equation}
	In E step this lower bound is maximized w.r.t q, and in M step, this lower bound is maximized w.r.t $R_3$. Optimal $q$ for E step is posterior probability:
	\begin{equation}\label{ELBO}
		\begin{aligned}
			q^{(t)} &= P(R_1|R_2, R_3^{(t-1)})\\
			&=P(\vec{\pi}_{1:K},\vec{Z}_{1:K}, Y|D_{1:K},\theta_{1:K}^{(t-1)},\vec{\alpha}^{(t-1)},B^{(t-1)})
		\end{aligned}
	\end{equation}
	\paragraph{Mean-field approximation}
	With mean field approximation, 
	\begin{equation}
		\begin{aligned}
			&q_{\Delta}(\vec{\pi}_{1:K},\vec{Z}_{1:K},Y|\gamma_{1:K},\vec{\Omega},w)\\
			=&\prod_{i}{q_1(\pi_i|\gamma_i)}{q_2(\vec{z}_{i}|\vec{\Omega}_{i},1)} \prod_{ij}{q_3(Y_{ij}|w_{ij})}
		\end{aligned}
	\end{equation}
	where $q_1$ is a Dirichlet, $q_2$ is a multinomial, $q_3$ is a Bernoulli, and $\Delta=(\gamma_{1:K},\vec{\Omega},w)$  represent the set of free
	variational parameters need to be estimated in the approximate distribution.
	
	With mean-field approximation, the expectaion of the lower bound can be calculated:
	
	\begin{equation}
		\begin{aligned}
			&L_{\Delta}:=\mathbb{E}_q \log P(R_1, R_2,R_3) - \log q(R_1) \\
			=&E_q \Bigg[\log
			\prod_{i=1}^k{\bigg(P(D_i|\theta_i)\prod_{j\neq i ,Y_{ij}=1}}{P(D_j|\theta_i)}\bigg) \exp(-\frac{\lambda}{2}\sum_{i}||\theta_i||^2)\\
			&\bigg(\prod_{ij}{P(Y_{ij}|\vec{z}_{i},\vec{z}_{j},B)} \prod_i{P(\vec{\pi}_i|\vec{\alpha}_i)P(\vec{z}_i|\vec{\pi}_i)}\bigg)-\\
			&\log \prod_{i}{q_1(\pi_i|\gamma_i)}{q_2(\vec{z}_{i}|\vec{\Omega}_{i},1)} \prod_{ij}{q_3(Y_{ij}|w_{ij})}\Bigg] \\
			=&E_q \Bigg[
			\sum_{i=1}^k{\bigg(\log P(D_i|\theta_i)+\sum_{j\neq i}}{Y_{ij}\log P(D_j|\theta_i)}\bigg)
			-\frac{\lambda}{2}\sum_{i}||\theta_i||^2\\
			&+\sum_{i,j}
			Y_{ij}\log\sum_{g,h} \vec{z}_{i,g} B(g,h)\vec{z}_{j,h}
			+\sum_{i,j}(1-Y_{ij})\log(1-\sum_{g,h} \vec{z}_{i,g} B(g,h)\vec{z}_{j,h})\\
			&+\sum_{i}\sum_g \vec{z}_{i,g}\log \vec{\pi}_{i,g}\\
			&+\sum_i \sum_g (\alpha_{i,g}-1)\log \vec{\pi}_{i,g}
			-\sum_i \sum_g \log \Gamma(\vec{\alpha}_{i,g})
			+\sum_i
			\log \Gamma(\sum_g\vec{\alpha}_{i,g})\\
			&-\sum_i \sum_g (\gamma_{i,g}-1)\log \vec{\pi}_{i,g}
			+\sum_i \sum_g \log \Gamma(\gamma_{i,g})
			-\sum_i
			\log \Gamma(\sum_g \gamma_{i,g})\\
			&-\sum_{i}\sum_g \vec{z}_{i,g}\log \vec{\Omega}_{i,g}\\
			&-\sum_{i,j}Y_{ij}\log w_{ij} 
			-\sum_{i,j}(1-Y_{ij})\log (1-w_{ij})\Bigg] \\
			=&\sum_{i=1}^k{\bigg(\log P(D_i|\theta_i)+\sum_{j\neq i}}{w_{ij}\log P(D_j|\theta_i)}\bigg)
			-\frac{\lambda}{2}\sum_{i}||\theta_i||^2\\
			&+\sum_{i,j}
			w_{ij}\sum_{g,h} \vec{\Omega}_{i,g}\vec{\Omega}_{j,h}\log B(g,h)
			+\sum_{i,j}
			(1-w_{ij})\sum_{g,h} \vec{\Omega}_{i,g}\vec{\Omega}_{j,h}\log (1-B(g,h))\\
			&+\sum_{i}\sum_g \vec{\Omega}_{i,g}\bigg(\psi(\gamma_{i,g})-\psi(\sum_k \gamma_{i,k})\bigg)\\
			&+\sum_i \sum_g (\vec{\alpha}_{g}-1)\bigg(\psi(\gamma_{i,g})-\psi(\sum_k \gamma_{i,k})\bigg)
			-\sum_i \sum_g \log \Gamma(\vec{\alpha}_{g})
			+\sum_i
			\log \Gamma(\sum_g\vec{\alpha}_{g})\\
			&-\sum_i \sum_g (\gamma_{i,g}-1)\bigg(\psi(\gamma_{i,g})-\psi(\sum_k \gamma_{i,k})\bigg)
			+\sum_i \sum_g \log \Gamma(\gamma_{i,g})
			-\sum_i
			\log \Gamma(\sum_g \gamma_{i,g})\\
			&-\sum_{i}\sum_g \vec{\Omega}_{i,g}\log \vec{\Omega}_{i,g}\\
			&-\sum_{i,j}w_{ij}\log w_{ij} 
			-\sum_{i,j}(1-w_{ij})\log (1-w_{ij})
		\end{aligned}
	\end{equation}\label{eq: Expectation ELBO}
	, where $\psi(x)$ is the digamma function defined as the logarithmic derivative of the gamma function: $\psi(x)=\frac{d \log \Gamma(x)}{d x}$.
	\paragraph{E step}
	Maximizing equation (\ref{eq: Expectation ELBO}) w.r.t. variational parameters $\Delta=(w,\gamma_{1:K},\vec{\Omega}_{\rightarrow},\vec{\Omega}_{\leftarrow})$. 
	\begin{itemize}
		\item $w_{ij}$:
		Setting $\nabla_{w} L_{\Delta}=0$:
		\begin{equation}
			\begin{aligned}
				&\log P(D_j|\theta_i)
				+
				\sum_{g,h} \vec{\Omega}_{i,g}\vec{\Omega}_{j,h}\log B(g,h)\\
				&-\sum_{g,h} \vec{\Omega}_{i,g}\vec{\Omega}_{j,h}\log (1-B(g,h))
				-1-\log w_{ij} 
				+1+\log (1-w_{ij}) = 0
			\end{aligned}
		\end{equation}
		Denote 
		\begin{equation}
			\begin{aligned}
				\hat{w_{ij}}=&\log P(D_j|\theta_i)
				+
				\sum_{g,h} \vec{\Omega}_{i,g}\vec{\Omega}_{j,h}\log B(g,h)\\
				&-\sum_{g,h} \vec{\Omega}_{i,g}\vec{\Omega}_{j,h}\log (1-B(g,h))
			\end{aligned}
		\end{equation}
		, then
		\begin{equation}
			\begin{aligned}
				w_{ij}^* = \frac{1}{1+\exp(-\hat{w_{ij}})}
				= \text{sigmoid}(\hat{w_{ij}})
			\end{aligned}
		\end{equation}
		
		\item $\gamma_{1:K}$: 
		Setting $\nabla_{\gamma_{i,g}} L_{\Delta}=0$:
		\begin{equation}
			\begin{aligned}
				&\bigg( \vec{\Omega}_{i,g}
				+\vec{\alpha}_{g}
				-\gamma_{i,g}
				\bigg)
				\psi^{'}(\gamma_{i,g})-
				\sum_t \bigg( \vec{\Omega}_{i,t}
				+\vec{\alpha}_{t}
				-\gamma_{i,t}\bigg)
				\psi^{'}(\sum_k \gamma_{i,k}) = 0
			\end{aligned}
		\end{equation}
		Therefore,
		\begin{equation}
			\begin{aligned}
				\gamma_{i,g}^* =  \vec{\Omega}_{i,g}
				+\vec{\alpha}_{g}
			\end{aligned}
\end{equation}
		
		\item $\vec{\Omega}_{i}$:
		Adding Lagrange multipliers into $L_{\Delta}$, and setting $\nabla_{\vec{\Omega}_{i,k}} \bigg(L_{\Delta}+\sum_{i}\lambda_{i} (\sum_h \vec{\Omega}_{i,h}-1)\bigg)=0$:
		\begin{equation}
			\begin{aligned}
				&\sum_j w_{ij} \sum_h \vec\Omega_{j, h} \log B(k, h) +\sum_{j}w_{ji}\sum_h \vec\Omega_{j, h} \log B(h, k) \\
				&\sum_j (1-w_{ij}) \sum_h \vec\Omega_{j, h} \log (1-B(k, h)) +\sum_{j}(1-w_{ji})\sum_h \vec\Omega_{j, h} \log (1-B(h, k))
				\\
				&+\psi(\gamma_{i,k})-\psi(\sum_g \gamma_{i,g})-1-\log \vec{\Omega}_{i,k} + \lambda_{i} = 0
			\end{aligned}
		\end{equation}
		Therefore, 
		\begin{equation}
			\begin{aligned}
				&\vec{\Omega}_{i,k}^*\propto 
				\exp \bigg(\sum_j w_{ij} \sum_h \vec\Omega_{j, h} \log B(k, h) +\sum_{j}w_{ji}\sum_h \vec\Omega_{j, h} \log B(h, k) \\
				&\sum_j (1-w_{ij}) \sum_h \vec\Omega_{j, h} \log (1-B(k, h)) +\sum_{j}(1-w_{ji})\sum_h \vec\Omega_{j, h} \log (1-B(h, k))
				\\
				&+\psi(\gamma_{i,k})-\psi(\sum_g \gamma_{i,g})\bigg)
			\end{aligned}
		\end{equation}
		We normalize $\vec{\Omega}_{i,k}^*$ to satisfy $\sum_k\vec{\Omega}_{i,k}^*=1$.
	\end{itemize}
	\paragraph{M step} We maximize the lower bound w.r.t. $\theta_{1:K},\vec{\alpha},B$.
	\begin{itemize}
		\item $\theta_{1:K}$: Using stochastic gradient ascent method,
		
		\begin{equation}
			\begin{aligned}
				&\theta_i \leftarrow \theta_i + \eta_1 \nabla_{\theta_i}L_{\Delta}\\
				&=\theta_i - \eta_1 \bigg(  
				\nabla_{\theta_i}L(D_i;\theta_i)+\sum_{j\neq i}w_{ij}\nabla_{\theta_i}L(D_j;\theta_i)
				+\lambda\theta_i
				\bigg)
			\end{aligned}
		\end{equation}
		
		\item $\vec{\alpha}$: Using gradient ascent method,
		\begin{equation}
			\begin{aligned}
				\vec{\alpha_{g}}\leftarrow
				\vec{\alpha_{g}}+\eta_2\bigg(
				\sum_i \bigg(\psi(\gamma_{i,g})-\psi(\sum_k \gamma_{i,k})\bigg)
				-K\sum_g \psi(\vec{\alpha}_{g})
				+K
				\psi(\sum_k\vec{\alpha}_{k})
				\bigg)
			\end{aligned}
		\end{equation}
		
		\item $B$: 
		Setting $\nabla_{B(g,h)} L_{\Delta}=0$:
		\begin{equation}
			\begin{aligned}
				\sum_{i,j}
				w_{ij}\vec{\Omega}_{i,g}\vec{\Omega}_{j,h}\frac{1}{B(g,h)} 
				-\sum_{i,j}
				(1-w_{ij}) \vec{\Omega}_{i,g}\vec{\Omega}_{j,h}\frac{1}{1-B(g,h)}=0
			\end{aligned}
		\end{equation}
		Therefore,
		\begin{equation}
			\begin{aligned}
				B(g,h)^*=\frac{\sum_{i,j}
					w_{ij}\vec{\Omega}_{i,g}\vec{\Omega}_{j,h}}{\sum_{i,j}
					\vec{\Omega}_{i,g}\vec{\Omega}_{j,h}}
			\end{aligned}
		\end{equation}
	\end{itemize}

	\section{SCooL-attention derivation}\label{sec:SCooL-att-EM}
	\begin{equation}
		\begin{aligned}
			&\text{($Y_i$ is a one-hot vector for client i, drawn from categorical distribution.)}\\
			&p_{ij}(\phi,\theta)=\text{softmax}(f(\theta_i;\phi)^Tf(\theta_j;\phi))\\
			&\vec{Y_i} \sim \text{Categorical}(p_{i1}(\phi,\theta),p_{i2}(\phi,\theta),...,p_{ik}(\phi,\theta))\\
			&P(Y|\theta) = \prod_{ij} p_{ij}(\phi,\theta)^{Y_{ij}} \\
			&P(\theta_{1:K})\propto \exp(-\frac{\lambda}{2}\sum_{i}||\theta_i||^2)\\
			&P(D_{1:K}|\theta_{1:K},Y)=
			\prod_{i=1}^K P(D_{1:K}|\theta_i,Y)=
			\prod_{i=1}^k{\bigg(P(D_i|\theta_i)\prod_{j\neq i ,Y_{ij}=1}}{P(D_j|\theta_i)}\bigg)
		\end{aligned}
	\end{equation}
	
	\subsection{Objective}
	Our final objective is:
	\begin{equation}
		\begin{aligned}
			&P(\theta_{1:K}|D_{1:K}) \propto P(\theta_{1:K},D_{1:K}) \\
			= &\int P(D_{1:K}|\theta_{1:K},Y)P(\theta_{1:K})P(Y|\theta_{1:K})dY \\
			= &\int
			\prod_{i=1}^k{\bigg(P(D_i|\theta_i)\prod_{j\neq i ,Y_{ij}=1}}{P(D_j|\theta_i)}\bigg)\exp(-\frac{\lambda}{2}\sum_{i}||\theta_i||^2) 
			\prod_{ij}p_{ij}(\phi,\theta)^{Y_{ij}}
			dY
		\end{aligned}
	\end{equation}
	\subsection{Optimization}
	\paragraph{ELBO}
	By Jensen's inequality, 
	\begin{equation}
		\begin{aligned}
			\log P(\theta_{1:K},D_{1:K})&
			=\log \int P(\theta_{1:K},D_{1:K},Y)d Y \\
			&=\log \int q(Y)\frac{P(\theta_{1:K},D_{1:K},Y)}{q(Y)}dY \\
			&\geq \int q(Y) \log \frac{P(\theta_{1:K},D_{1:K},Y)}{q(Y)}dY \\
			&=\mathbb{E}_q \log P(\theta_{1:K},D_{1:K},Y)) - \log q(Y) =: \mathbb{L}_{\Delta}
		\end{aligned}
	\end{equation}
	In E step this lower bound is maximized w.r.t q, and in M step, this lower bound is maximized w.r.t $R_3$. Optimal $q$ for E step is posterior probability:
	\begin{equation}\label{ELBO}
		\begin{aligned}
			q^{(t)} &= P(Y|\theta_{1:K}^{(t-1)})
		\end{aligned}
	\end{equation}
	\paragraph{Mean-field approximation}
	With mean field approximation, 
	\begin{equation}
		\begin{aligned}
			&q_{\Delta}(Y)=\prod_{i}q_{\Delta}(\vec{Y_i})
			=\text{Categorical}(p_{i1},p_{i2},...,p_{iK})
		\end{aligned}
	\end{equation}
	
	With mean-field approximation, the expectaion of the lower bound can be calculated:
	
	\begin{equation}
		\begin{aligned}
			&L_{\Delta}:=\mathbb{E}_q \log P(\theta_{1:K},D_{1:K},Y) - \log q(Y) \\
			=&E_q \bigg[\log
			\prod_{i=1}^k{\bigg(P(D_i|\theta_i)\prod_{j\neq i ,Y_{ij}=1}}{P(D_j|\theta_i)}\bigg)\exp(-\frac{\lambda}{2}\sum_{i}||\theta_i||^2) \\
			&\prod_{ij}p_{ij}(\phi,\theta)^{Y_{ij}} - \log \prod_{ij}w_{ij}^{Y_{ij}}
			\bigg] \\
			=&E_q \bigg[
			\sum_{i=1}^k{\bigg(\log P(D_i|\theta_i)+\sum_{j\neq i}}{Y_{ij}\log P(D_j|\theta_i)}\bigg)
			-\frac{\lambda}{2}\sum_{i}||\theta_i||^2\\
			&+\sum_{ij}Y_{ij}\log p_{ij}(\phi,\theta)-\sum_{ij}Y_{ij}\log w_{ij}\bigg]\\
			=&\sum_{i=1}^k{\bigg(\log P(D_i|\theta_i)+\sum_{j\neq i}}{w_{ij}\log P(D_j|\theta_i)}\bigg)
			-\frac{\lambda}{2}\sum_{i}||\theta_i||^2\\
			&+\sum_{ij}w_{ij}\log p_{ij}(\phi,\theta)-\sum_{ij}w_{ij}\log w_{ij}
		\end{aligned}
	\end{equation}
	\paragraph{E step}
	\begin{itemize}
		\item $w_{ij}$:
		Setting $\nabla_{w_{ij}} L_{\Delta}=0$:
		\begin{equation}
			\begin{aligned}
				\log P(D_j|\theta_i)
				+\log p_{ij}(\phi,\theta)-1-\log w_{ij}=0
			\end{aligned}
		\end{equation}
	\end{itemize}
	Then,
	\begin{equation}
		\begin{aligned}
			w_{ij}^* \propto 
			\text{exp}\bigg(
			\log P(D_j|\theta_i)
			+\log p_{ij}(\phi,\theta)
			\bigg)
		\end{aligned}
	\end{equation}
	We need to further normalize $w_{ij}$ to satisfy $\sum_j w_{ij}=1$, i.e. $w_{ij}=\frac{w_{ij}^*}{\sum_j w_{ij}^*}$.
	
	\paragraph{M step}
	\begin{itemize}
		\item $\theta_{1:K}$:
		We use stochastic gradient descent to optimize it:
		\begin{equation}
			\begin{aligned}
				&\theta_i \leftarrow \theta_i + \eta_1 \nabla_{\theta_i}L_{\Delta}\\
				&=\theta_i - \eta_1 \bigg(  
				\nabla_{\theta_i}L(D_i;\theta_i)+\sum_{j\neq i}w_{ij}\nabla_{\theta_i}L(D_j;\theta_i)
				+\lambda\theta_i
				-\sum_{ij}w_{ij}\nabla_{\theta_i}\log p_{ij}(\phi,\theta)\\
				&=\theta_i - \eta_1 \bigg(  
				\nabla_{\theta_i}L(D_i;\theta_i)+\sum_{j\neq i}w_{ij}\nabla_{\theta_i}L(D_j;\theta_i)
				+\lambda\theta_i\\
				&+\sum_{ij}\nabla_{\theta_i}\text{cross-entropy}(\vec{w_i},\vec{p_{i}}(\phi,\theta))
				\bigg)
			\end{aligned}
		\end{equation}
		\item $\phi$: We use stochastic gradient descent to optimize it:
		\begin{equation}
			\begin{aligned}
				&\phi \leftarrow \phi + \eta_2 \nabla_{\phi}L_{\Delta}\\
				&=\phi + \eta_2 \nabla_{\phi}
				\bigg(\sum_{ij}w_{ij}\log p_{ij}(\phi)\bigg)\\
				&=\phi - \eta_2 \sum_i\nabla_{\phi}
				\text{cross-entropy}(\vec{w_i},\vec{p_{i}}(\phi,\theta))
			\end{aligned}
		\end{equation}
	\end{itemize}

\section{SCooL-MMSBM}
We present an additional instantiation of SCooL framework,  which we use mixed membership stochastic blockmodels (MMSBM) as the prior, allowing each user to simultaneously cooperate with multiple groups of users with different probability, thereby capturing more complex cooperation between users' tasks.
\begin{equation}
\begin{aligned}
    P(\theta_{1:K}|D_{1:K}) \propto P(\theta_{1:K},D_{1:K}) \\= \int P(D_{1:K}|\theta_{1:K},Y)P(\theta_{1:K}|Y)P(Y)dY
\end{aligned}
\end{equation}
with
\begin{equation}
    \begin{aligned}
        &P(\theta_{1:K},Y)=P(\theta_{1:K})P(Y)\\
        &P(Y) =\int P(Y,\vec{\pi}_{1:K},Z_{\rightarrow},Z_{\leftarrow}|\vec{\alpha},B)d(\vec{\pi}_{1:K},Z_{\rightarrow},Z_{\leftarrow}) \\
        &P(\theta_{1:K})\propto \exp(-\frac{\lambda}{2}\sum_{i}||\theta_i||^2)\\
        &P(D_{1:K}|\theta_{1:K},Y)=
        \prod_{i=1}^K P(D_{1:K}|\theta_i,Y)=
        \prod_{i=1}^k{\bigg(P(D_i|\theta_i)\prod_{j\neq i ,Y_{ij}=1}}{P(D_j|\theta_i)}\bigg)
    \end{aligned}
    \end{equation}
\subsubsection{Objective}
Modelling $P(Y)$ as MMSBM:
\begin{itemize}
	\item For each agent $i \in K$:
	\begin{itemize}
		\item Draw a $M$ dimensional mixed membership vector $\vec{\pi}_i \sim
		\textrm{Dirichlet}( \vec\alpha )$.
	\end{itemize}
	\item For each pair of agents $(i,j) \in K \times K$:
	\begin{itemize}
		\item Draw membership indicator for the initiator, $\vec z_{i
			\rightarrow j} ~ \sim {\rm Multinomial}(\vec\pi_{i})$.
		\item Draw membership indicator for the receiver, $\vec z_{i
			\leftarrow j} ~ \sim {\rm Multinomial}(\vec\pi_{j})$.
		\item Sample the value of their interaction, $Y_{ij} \sim {\rm
			Bernoulli}(\vec z_{i
			\rightarrow j}^{\ T} B ~ \vec z_{i
			\leftarrow j})$.
	\end{itemize}
\end{itemize}

Under the MMSBM, the marginal distribution of $Y$ is:
\begin{equation}
	\begin{aligned}
		&P(Y|\vec{\alpha},B)=\int P(Y,\vec{\pi}_{1:K},Z_{\rightarrow},Z_{\leftarrow}|\vec{\alpha},B)d(\vec{\pi}_{1:K},Z_{\rightarrow},Z_{\leftarrow})\\
		&= \int \bigg(\prod_{ij}{P(Y_{ij}|\vec{z}_{i\rightarrow j},\vec{z}_{i\leftarrow j},B)P(\vec{z}_{i \rightarrow j}|\vec{\pi}_i)P(\vec{z}_{i \leftarrow j}|\vec{\pi_j})}\\
		&\ \ \ \ \ \ \ \ \ \  \prod_i{P(\vec{\pi}_i|\vec{\alpha}_i)}\bigg)d(\vec{\pi}_{1:K},Z_{\rightarrow},Z_{\leftarrow}) 
	\end{aligned}
\end{equation}

Our final objective is:
\begin{equation}
\begin{aligned}
    &P(\theta_{1:K}|D_{1:K}) \propto P(\theta_{1:K},D_{1:K}) \\
    = &\int P(D_{1:K}|\theta_{1:K},Y)P(\theta_{1:K})P(Y)dY \\
    = &\int
        \prod_{i=1}^k{\bigg(P(D_i|\theta_i)\prod_{j\neq i ,Y_{ij}=1}}{P(D_j|\theta_i)}\bigg) \exp(-\frac{\lambda}{2}\sum_{i}||\theta_i||^2)\\
        &\bigg(\prod_{ij}{P(Y_{ij}|\vec{z}_{i\rightarrow j},\vec{z}_{i\leftarrow j},B)P(\vec{z}_{i \rightarrow j}|\vec{\pi}_i)P(\vec{z}_{i \leftarrow j}|\vec{\pi_j})} \prod_i{P(\vec{\pi}_i|\vec{\alpha}_i)}\bigg)d(\vec{\pi}_{1:K},Z_{\rightarrow},Z_{\leftarrow},Y)
\end{aligned}
\end{equation}
\subsubsection{Optimization}
\paragraph{ELBO}
Rewrite $R_1 = (Y,\vec{\pi}_{1:K},Z_{\rightarrow},Z_{\leftarrow})$, $R_2 = (D_{1:K})$, $R_3=(\theta_{1:K},\vec{\alpha},B)$.
\begin{equation}
	\begin{aligned}
		\log P(R_2,R_3)&
		=\log \int P(R_1, R_2,R_3)d R_1 \\
		&=\log \int q(R_1)\frac{P(R_1, R_2,R_3)}{q(R_1)}dR_1 \\
		&\geq \int q(R_1) \log \frac{P(R_1, R_2,R_3)}{q(R_1)}dR_1 \\
		&=\mathbb{E}_q \log P(R_1, R_2,R_3) - \log q(R_1) =: \mathbb{L}(q,R_3)
	\end{aligned}
\end{equation}
In E step this lower bound is maximized w.r.t q, and in M step, this lower bound is maximized w.r.t $R_3$. Optimal $q$ for E step is posterior probability:
\begin{equation}\label{ELBO}
	\begin{aligned}
		q^{(t)} &= P(R_1|R_2, R_3^{(t-1)})\\
		&=P(Y,\vec{\pi}_{1:K},Z_{\rightarrow},Z_{\leftarrow}|D_{1:K},\theta_{1:K}^{(t-1)},\vec{\alpha}^{(t-1)},B^{(t-1)})
	\end{aligned}
\end{equation}
\paragraph{Mean-field approximation}
With mean field approximation, 
\begin{equation}
	\begin{aligned}
		&q_{\Delta}(Y,\vec{\pi}_{1:K},Z_{\rightarrow},Z_{\leftarrow}|w,\gamma_{1:K},\vec{\phi}_{\rightarrow},\vec{\phi}_{\leftarrow})\\
		=&\prod_{i}{q_1(\pi_i|\gamma_i)}\prod_{i,j}{q_2(\vec{z}_{i\rightarrow j}|\vec{\phi}_{i\rightarrow j},1)q_2(\vec{z}_{i\leftarrow j}|\vec{\phi}_{i\leftarrow j},1)} \prod_{ij}{q_3(Y_{ij}|w_{ij})}
	\end{aligned}
\end{equation}
where $q_1$ is a Dirichlet, $q_2$ is a multinomial, $q_3$ is a Bernoulli, and $\Delta=(w,\gamma_{1:K},\vec{\phi}_{\rightarrow},\vec{\phi}_{\leftarrow})$  represent the set of free
variational parameters need to be estimated in the approximate distribution.

With mean-field approximation, the expectaion of the lower bound can be calculated:

\allowdisplaybreaks[4]
\begin{align*}
    &L_{\Delta}:=\mathbb{E}_q \log P(R_1, R_2,R_3) - \log q(R_1) \\
    =&E_q \Bigg[\log
        \prod_{i=1}^k{\bigg(P(D_i|\theta_i)\prod_{j\neq i ,Y_{ij}=1}}{P(D_j|\theta_i)}\bigg)\exp(-\frac{\lambda}{2}\sum_{i}||\theta_i||^2) \\
        &\bigg(\prod_{ij}{P(Y_{ij}|\vec{z}_{i\rightarrow j},\vec{z}_{i\leftarrow j},B)P(\vec{z}_{i \rightarrow j}|\vec{\pi}_i)P(\vec{z}_{i \leftarrow j}|\vec{\pi_j})} \prod_i{P(\vec{\pi}_i|\vec{\alpha}_i)}\bigg)-
        \displaybreak[4]
        \\
        &\log \prod_{i}{q_1(\pi_i|\gamma_i)}\prod_{i,j}{q_2(\vec{z}_{i\rightarrow j}|\vec{\phi}_{i\rightarrow j},1)q_2(\vec{z}_{i\leftarrow j}|\vec{\phi}_{i\leftarrow j},1)} \prod_{ij}{q_3(Y_{ij}|w_{ij})}\Bigg] 
        \\
        =&E_q \Bigg[
        \sum_{i=1}^k{\bigg(\log P(D_i|\theta_i)+\sum_{j\neq i}}{Y_{ij}\log P(D_j|\theta_i)}\bigg)
        -\frac{\lambda}{2}\sum_{i}||\theta_i||^2\\
        &+\sum_{i,j}
        Y_{ij}\log\sum_{g,h} \vec{z}_{i\rightarrow j,g} B(g,h)\vec{z}_{i\leftarrow j,h}
        +\sum_{i,j}(1-Y_{ij})\log(1-\sum_{g,h} \vec{z}_{i\rightarrow j,g} B(g,h)\vec{z}_{i\leftarrow j,h})
        \\
        &+\sum_{i,j}\sum_g \vec{z}_{i \rightarrow j,g}\log \vec{\pi}_{i,g}
        +\sum_{i,j}\sum_g \vec{z}_{i \leftarrow j,g}\log \vec{\pi}_{j,g}\\
        &+\sum_i \sum_g (\alpha_{i,g}-1)\log \vec{\pi}_{i,g}
        -\sum_i \sum_g \log \Gamma(\vec{\alpha}_{i,g})
        +\sum_i
        \log \Gamma(\sum_g\vec{\alpha}_{i,g})\\
        &-\sum_i \sum_g (\gamma_{i,g}-1)\log \vec{\pi}_{i,g}
        +\sum_i \sum_g \log \Gamma(\gamma_{i,g})
        -\sum_i
        \log \Gamma(\sum_g \gamma_{i,g})\\
        &-\sum_{i,j}\sum_g \vec{z}_{i \rightarrow j,g}\log \vec{\phi}_{i\rightarrow j,g}
        -\sum_{i,j}\sum_g \vec{z}_{i \leftarrow j,g}\log \vec{\phi}_{i \leftarrow j,g}\\
        &-\sum_{i,j}Y_{ij}\log w_{ij} 
        -\sum_{i,j}(1-Y_{ij})\log (1-w_{ij})\Bigg] \\
        =&\sum_{i=1}^k{\bigg(\log P(D_i|\theta_i)+\sum_{j\neq i}}{w_{ij}\log P(D_j|\theta_i)}\bigg)
        -\frac{\lambda}{2}\sum_{i}||\theta_i||^2\\
        &+\sum_{i,j}
        w_{ij}\sum_{g,h} \vec{\phi}_{i\rightarrow j,g}\vec{\phi}_{i\leftarrow j,h}\log B(g,h)
        +\sum_{i,j}
        (1-w_{ij})\sum_{g,h} \vec{\phi}_{i\rightarrow j,g}\vec{\phi}_{i\leftarrow j,h}\log (1-B(g,h))\\
        &+\sum_{i,j}\sum_g \vec{\phi}_{i \rightarrow j,g}\bigg(\psi(\gamma_{i,g})-\psi(\sum_k \gamma_{i,k})\bigg)
        +\sum_{i,j}\sum_g \vec{\phi}_{i \leftarrow j,g}\bigg(\psi(\gamma_{j,g})-\psi(\sum_k \gamma_{j,k})\bigg)\\
        &+\sum_i \sum_g (\vec{\alpha}_{g}-1)\bigg(\psi(\gamma_{i,g})-\psi(\sum_k \gamma_{i,k})\bigg)
        -\sum_i \sum_g \log \Gamma(\vec{\alpha}_{g})
        +\sum_i
        \log \Gamma(\sum_g\vec{\alpha}_{g})\\
        &-\sum_i \sum_g (\gamma_{i,g}-1)\bigg(\psi(\gamma_{i,g})-\psi(\sum_k \gamma_{i,k})\bigg)
        +\sum_i \sum_g \log \Gamma(\gamma_{i,g})
        -\sum_i
        \log \Gamma(\sum_g \gamma_{i,g})\\
        &-\sum_{i,j}\sum_g \vec{\phi}_{i \rightarrow j,g}\log \vec{\phi}_{i\rightarrow j,g}
        -\sum_{i,j}\sum_g \vec{\phi}_{i \leftarrow j,g}\log \vec{\phi}_{i \leftarrow j,g}\\
        &-\sum_{i,j}w_{ij}\log w_{ij} 
        -\sum_{i,j}(1-w_{ij})\log (1-w_{ij})
\end{align*}
\label{eq: Expectation ELBO}
, where $\psi(x)$ is the digamma function defined as the logarithmic derivative of the gamma function: $\psi(x)=\frac{d \log \Gamma(x)}{d x}$.
\paragraph{E step}
Maximizing equation \ref{eq: Expectation ELBO} w.r.t. variational parameters $\Delta=(w,\gamma_{1:K},\vec{\phi}_{\rightarrow},\vec{\phi}_{\leftarrow})$. 
\begin{itemize}
    \item $w_{ij}$:
    Setting $\nabla_{w} L_{\Delta}=0$:
\begin{equation}
    \begin{aligned}
        &\log P(D_j|\theta_i)
        +
        \sum_{g,h} \vec{\phi}_{i\rightarrow j,g}\vec{\phi}_{i\leftarrow j,h}\log B(g,h)\\
        &-\sum_{g,h} \vec{\phi}_{i\rightarrow j,g}\vec{\phi}_{i\leftarrow j,h}\log (1-B(g,h))
        -1-\log w_{ij} 
        +1+\log (1-w_{ij}) = 0
    \end{aligned}
\end{equation}
Denote 
\begin{equation}
    \begin{aligned}
        \hat{w_{ij}}=&\log P(D_j|\theta_i)
        +
        \sum_{g,h} \vec{\phi}_{i\rightarrow j,g}\vec{\phi}_{i\leftarrow j,h}\log B(g,h)\\
        &-\sum_{g,h} \vec{\phi}_{i\rightarrow j,g}\vec{\phi}_{i\leftarrow j,h}\log (1-B(g,h))
    \end{aligned}
\end{equation}
, then
\begin{equation}
    \begin{aligned}
        w_{ij}^* = \frac{1}{1+\exp(-\hat{w_{ij}})}
        = \text{sigmoid}(\hat{w_{ij}})
    \end{aligned}
\end{equation}

\item $\gamma_{1:K}$: 
Setting $\nabla_{\gamma_{i,g}} L_{\Delta}=0$:
\begin{equation}
    \begin{aligned}
    &\sum_{j} \vec{\phi}_{i \rightarrow j,g}\bigg(\psi^{'}(\gamma_{i,g})-\psi^{'}(\sum_k \gamma_{i,k})\bigg)
        +\sum_{j} \vec{\phi}_{j \leftarrow i,g}\bigg(\psi^{'}(\gamma_{i,g})-\psi^{'}(\sum_k \gamma_{i,k})\bigg)\\
    &+(\vec{\alpha}_{g}-1)\bigg(\psi^{'}(\gamma_{i,g})-\psi^{'}(\sum_k \gamma_{i,k})\bigg)
    -\bigg(\psi(\gamma_{i,g})-\psi(\sum_k \gamma_{i,k})\bigg)\\
    &-(\gamma_{i,g}-1)\bigg(\psi^{'}(\gamma_{i,g})-\psi^{'}(\sum_k \gamma_{i,k})\bigg)
        +\psi(\gamma_{i,g})
        -\psi(\sum_k \gamma_{i,k})=0\\
        \\
        &\bigg(
        \sum_{j} \vec{\phi}_{i \rightarrow j,g}
        +\sum_{j} \vec{\phi}_{j \leftarrow i,g}
        +\vec{\alpha}_{g}
        -\gamma_{i,g}
        \bigg)
        \bigg(\psi^{'}(\gamma_{i,g})-\psi^{'}(\sum_k \gamma_{i,k})\bigg) = 0
    \end{aligned}
\end{equation}
Therefore,
\begin{equation}
    \begin{aligned}
        \gamma_{i,g}^* =  \vec{\alpha}_g + \sum_j \vec{\phi}_{i \rightarrow j,g} + \sum_j \vec{\phi}_{j \leftarrow i,g}
    \end{aligned}
\end{equation}

\item $\vec{\phi}_{i \rightarrow j}$:
Adding Lagrange multipliers into $L_{\Delta}$, and setting $\nabla_{\vec{\phi}_{i \rightarrow j,k}} \bigg(L_{\Delta}+\sum_{i,j}\lambda_{ij} (\sum_h \vec{\phi}_{i \rightarrow j,h}-1)\bigg)=0$:
\begin{equation}
    \begin{aligned}
        &w_{ij}\sum_h \vec{\phi}_{i \leftarrow j,h} \log B(g,h) + (1-w_{ij})\sum_h \vec{\phi}_{i \leftarrow j,h} \log (1-B(g,h)) \\
        &+\psi(\gamma_{i,k})-\psi(\sum_g \gamma_{i,g})-1-\log \vec{\phi}_{i \rightarrow j,k} + \lambda_{ij} = 0
    \end{aligned}
\end{equation}
Therefore, 
\begin{equation}
    \begin{aligned}
        &\vec{\phi}_{i \rightarrow j,k}^*\propto 
        \exp \bigg(w_{ij}\sum_h \vec{\phi}_{i \leftarrow j,h} \log B(k,h) + \\
        &(1-w_{ij})\sum_h \vec{\phi}_{i \leftarrow j,h} \log (1-B(k,h)) 
        +\psi(\gamma_{i,k})-\psi(\sum_g \gamma_{i,g})\bigg)
    \end{aligned}
\end{equation}
We normalize $\vec{\phi}_{i \rightarrow j,k}^*$ to satisfy $\sum_k\vec{\phi}_{i \rightarrow j,k}^*=1$.

\item $\vec{\phi}_{i \leftarrow j}$:
Following similar derivations as $\vec{\phi}_{i \rightarrow j}$, we get:
\begin{equation}
    \begin{aligned}
        &\vec{\phi}_{i \leftarrow j,h}^*\propto 
        \exp \bigg(w_{ij}\sum_g \vec{\phi}_{i \rightarrow j,g} \log B(g,h) + \\
        &(1-w_{ij})\sum_g \vec{\phi}_{i \rightarrow j,g} \log (1-B(g,h)) 
        +\psi(\gamma_{j,h})-\psi(\sum_g \gamma_{j,g})\bigg)
    \end{aligned}
\end{equation}
We normalize $\vec{\phi}_{i \leftarrow j,h}^*$ to satisfy $\sum_h\vec{\phi}_{i \leftarrow j,h}^*=1$.
\end{itemize}
\paragraph{M step} We maximize the lower bound w.r.t. $\theta_{1:K},\vec{\alpha},B$.
\begin{itemize}
    \item $\theta_{1:K}$: Using stochastic gradient ascent method,
    
    \begin{equation}
        \begin{aligned}
            &\theta_i \leftarrow \theta_i + \eta_1 \nabla_{\theta_i}L_{\Delta}\\
            &=\theta_i - \eta_1 \bigg(  
            \nabla_{\theta_i}L(D_i;\theta_i)+\sum_{j\neq i}w_{ij}\nabla_{\theta_i}L(D_j;\theta_i)
        +\lambda\theta_i
            \bigg)
        \end{aligned}
    \end{equation}

\item $\vec{\alpha}$: Using gradient ascent method,
\begin{equation}
    \begin{aligned}
        \vec{\alpha_{g}}\leftarrow
        \vec{\alpha_{g}}+\eta_2\bigg(
        \sum_i \bigg(\psi(\gamma_{i,g})-\psi(\sum_k \gamma_{i,k})\bigg)
        -K\sum_g \psi(\vec{\alpha}_{g})
        +K
        \psi(\sum_k\vec{\alpha}_{k})
        \bigg)
    \end{aligned}
\end{equation}

\item $B$: 
Setting $\nabla_{B(g,h)} L_{\Delta}=0$:
\begin{equation}
    \begin{aligned}
        \sum_{i,j}
        w_{ij}\vec{\phi}_{i\rightarrow j,g}\vec{\phi}_{i\leftarrow j,h}\frac{1}{B(g,h)} 
        -\sum_{i,j}
        (1-w_{ij}) \vec{\phi}_{i\rightarrow j,g}\vec{\phi}_{i\leftarrow j,h}\frac{1}{1-B(g,h)}=0
    \end{aligned}
\end{equation}
Therefore,
\begin{equation}
    \begin{aligned}
        B(g,h)^*=\frac{\sum_{i,j}
        w_{ij}\vec{\phi}_{i\rightarrow j,g}\vec{\phi}_{i\leftarrow j,h}}{\sum_{i,j}
        \vec{\phi}_{i\rightarrow j,g}\vec{\phi}_{i\leftarrow j,h}}
    \end{aligned}
\end{equation}
\end{itemize}
	

\end{document}